\documentclass{style}

\usepackage{wrapfig}
\usepackage{booktabs}
\usepackage{multirow}
\usepackage{subcaption}

\usepackage[table,xcdraw]{xcolor}

\title{
    \fontsize{20pt}{24pt}\sffamily\bfseries\color{black}
    Thinking with Programming Vision: \\
    Towards a Unified View for Thinking with Images
}
\author{ 
    \fontsize{13pt}{15.5pt}\selectfont\sffamily\color{black}
    Zirun Guo$^{1,2}$ \quad Minjie Hong$^{1}$ \quad Feng Zhang$^{2}$ \quad Kai Jia$^{2}$ \quad Tao Jin$^{1}$ \\[0.5em]
    \normalsize\sffamily\color{black} $^1$ Zhejiang University \quad
    \normalsize\sffamily\color{black} $^2$ ByteDance, BandAI \\
}

\email{zrguo.cs@gmail.com}
\project{https://github.com/ByteDance-BandAI/CodeVision}

\begin{document}

\maketitle
\thispagestyle{firstpage}

\begin{abstract}
Multimodal large language models (MLLMs) that ``think with images'' can interactively use tools to reason about visual inputs, but current approaches often rely on a narrow set of tools with limited real-world necessity and scalability. In this work, we first reveal a critical and previously overlooked weakness: even state-of-the-art MLLMs are surprisingly brittle, showing significant performance degradation on images with simple orientation changes or natural corruptions, underscoring the need for more robust tool-based reasoning. To address this, we propose \textit{CodeVision}, a flexible and scalable ``code-as-tool'' framework where the model generates code as a universal interface to invoke any image operation, moving beyond fixed tool registries. We train our model using a two-stage methodology, beginning with Supervised Fine-Tuning (SFT) on a high-quality dataset curated for complex, multi-turn tool composition and error recovery, followed by Reinforcement Learning (RL) with a novel and dense process reward function to encourage strategic and efficient tool use. To facilitate this research, we construct new SFT and RL datasets and introduce a challenging new benchmark suite designed to rigorously evaluate robustness to orientation changes and multi-tool reasoning. Experiments on Qwen2.5-VL and Qwen3-VL series show that our approach significantly improves model performance and fosters emergent capabilities such as flexible tool composition, efficient chained execution, and robust error recovery from runtime feedback.
\end{abstract}

\section{Introduction}

Thinking with images~\citep{o3} equips multimodal large language models (MLLMs) with interactive, tool-augmented reasoning over visual inputs. Rather than passively describing an image, the model actively manipulates it using tools (\textit{e.g.,} ``zoom in'', OCR) to acquire the evidence needed for reliable reasoning. Despite rapid progress, current research still faces three limitations:

\begin{enumerate}[label=\circled{\arabic*}]
    \item \textbf{The necessity of tools}: Current methods mainly emphasize the ``crop'' tool, which zooms in on regions of interest for clearer observation, and evaluate tool ability on benchmarks such as V*~\citep{wu2024v} and HRBench~\citep{wang2025divide}. Yet the benefit is marginal: using tools often yields only 2--5\% accuracy gains, and reinforcement learning (RL) without tools can match those results. This suggests that the potential---and necessity---of tools is not being fully exercised by existing tasks.
    \item \textbf{Flexibility and scalability}: Methods frequently require manually specifying tool names and arguments~\citep{su2025openthinkimg}, which is brittle and not scalable. Even renaming a tool (\textit{e.g.,} \texttt{crop} \(\rightarrow\) \texttt{zoomin}) can necessitate retraining, hindering generalization to new tools and argument schemas.
    \item \textbf{Multi-turn, multi-tool use}: Many systems support only a single tool or a few tools within a single turn. While several works~\citep{lai2025mini,zhang2025thyme} explore multi-turn settings, they largely focus on repeated cropping rather than composing different tools across multiple turns, which is what real-world tasks often require.
\end{enumerate}

\begin{wrapfigure}{r}{8.5cm}
    \centering
    \includegraphics[width=0.5\textwidth]{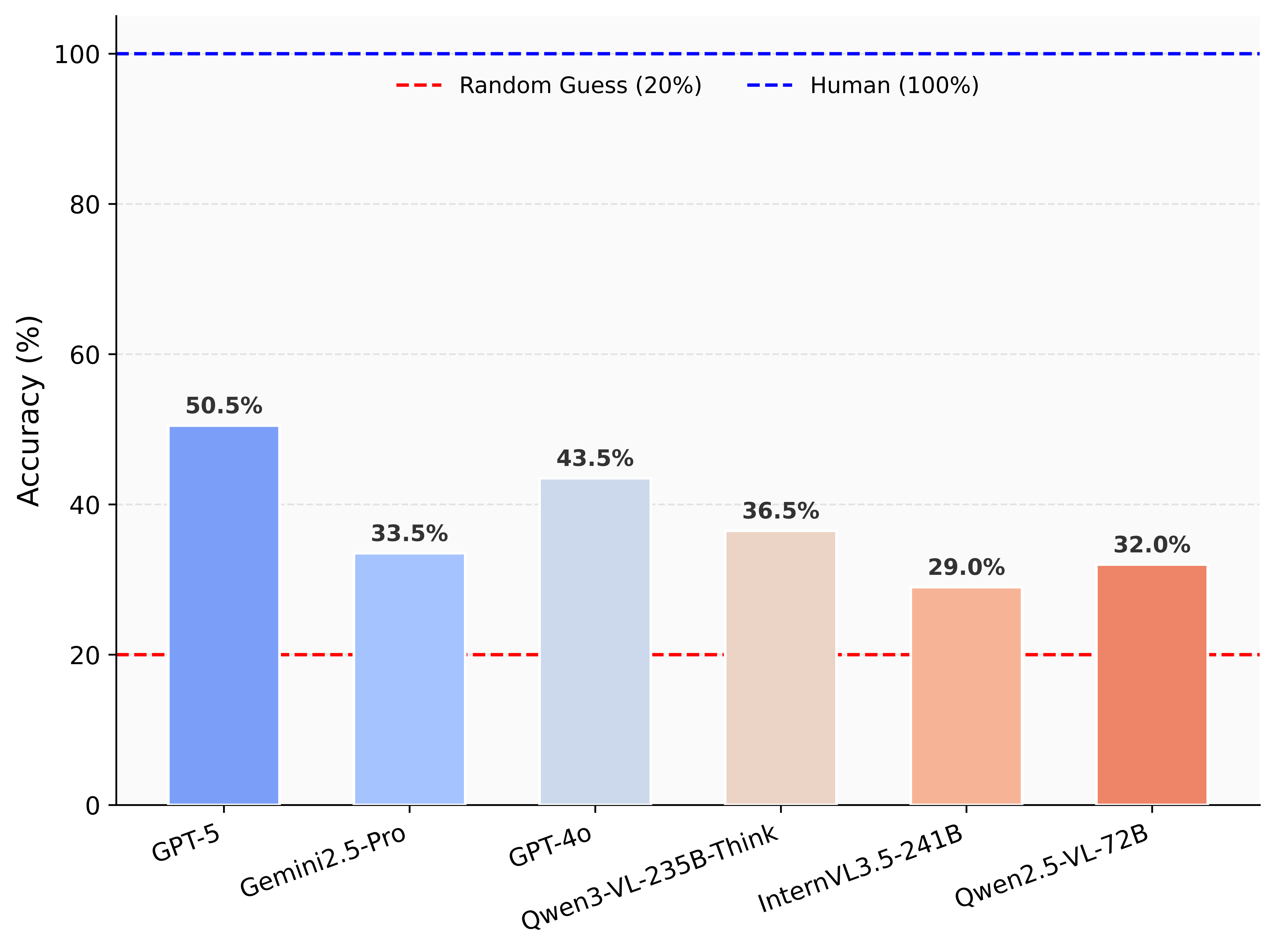}
    \caption{Diagnostic results on image orientation identification.}
    \label{fig:diagnostic}
    \vskip -0.1in
\end{wrapfigure}

Based on these observations, we propose a new framework to address all three challenges.

To create scenarios where tools are genuinely required, we examine ubiquitous real-world perturbations: incorrect image orientation due to landscape/portrait capture and mirrored selfies that confound text recognition. Compared with cropping, restoring the canonical orientation is often strictly more necessary for downstream recognition and reasoning. We conduct a simple diagnostic: starting from 200 images from various domains, we uniformly apply one of five transformations---rotation by 90/180/270 degrees, horizontal flip, or vertical flip---and ask models to identify the transformation (a five-way multiple-choice question). As illustrated in Figure~\ref{fig:diagnostic}, even state-of-the-art models like GPT-5~\citep{gpt5} and Gemini2.5 Pro~\citep{comanici2025gemini} perform poorly, whereas humans achieve 100\% accuracy easily. Remarkably, as summarized in Table~\ref{tab:main_results1}, simple rotation/flip operations can reduce model performance by up to 80\%, revealing substantial brittleness of current MLLMs and underscoring the necessity of using tools for these real-world corruptions.

\begin{wrapfigure}{r}{8.5cm}
    \centering
    \includegraphics[width=0.5\textwidth]{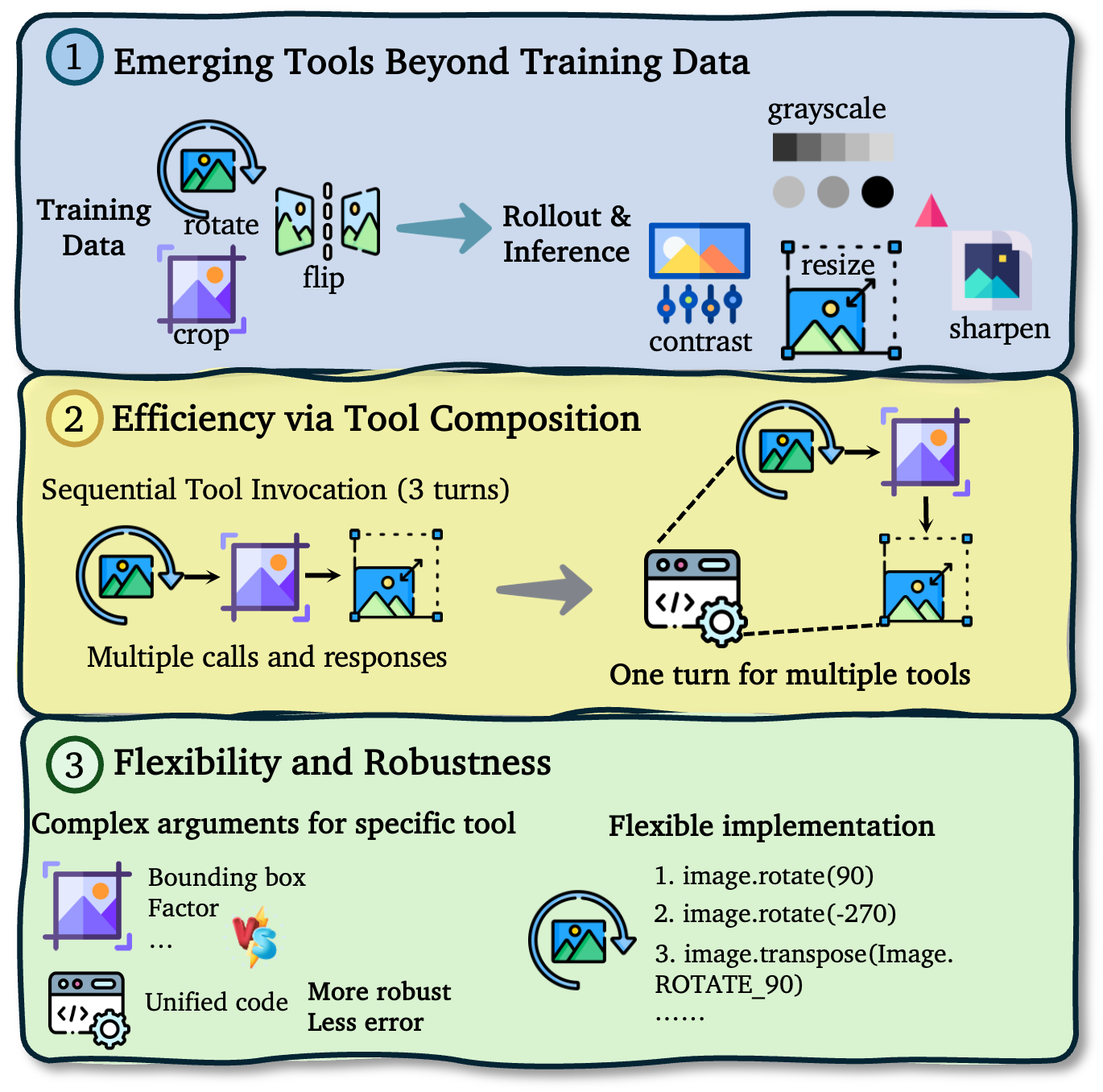}
    \caption{Three advantages of CodeVision we observe in the training and inference stage.}
    \label{fig:advantages}
    \vskip -0.1in
\end{wrapfigure}

To address the second challenge, inspired by OpenAI o3~\citep{o3}, we treat \emph{code} itself as a unified tool: the model writes code that invokes whatever image operations are needed. This eliminates hand-crafted tool name/argument specifications and dramatically improves generalization: the model is no longer restricted to a fixed registry but can call an effectively unbounded set of tools through code. While the paradigm of using code as a tool has been explored previously~\citep{chen2023program, zhang2025thyme}, we unlock three notable advantages through our principled RL training and rigorous dataset construction (see Figure~\ref{fig:advantages}): (1) \emph{Emergence of new tools}: the model calls tools that never appeared in the RL training data to solve novel problems; (2) \emph{Efficiency}: the model chains multiple tools within a single execution; (3) \emph{Robustness}: the model leverages runtime error messages and outputs to revise code, improving failure recovery and out-of-distribution generalization.

To target the third challenge, we design datasets and a benchmark that require composing multiple tools across multiple turns, and we introduce dense process rewards that encourage effective tool selection and steady progress throughout the dialogue. This setting better reflects realistic problem solving and promotes the development of models that plan, adapt, and utilize different tools across steps.
In summary, our contributions are threefold: 
\begin{itemize}
    \item We identify a critical brittleness in state-of-the-art MLLMs and propose a flexible ``code-as-tool'' framework that treats code as a universal interface for tool invocation, enhancing scalability and generalization.
    \item We construct high-quality SFT and RL datasets focused on multi-turn, multi-tool composition and error handling, and introduce three new benchmarks covering both single- and multi-tool scenarios to rigorously evaluate model robustness and complex tool use.
    \item Our experiments demonstrate that our approach significantly improves model performance on these challenging benchmarks and validates the benefits of the code-as-tool paradigm, which enables emergent tool use, efficient tool chaining, and robust error recovery.
\end{itemize}

\section{Related Work}
\textbf{Thinking with Images.} OpenAI's o3~\citep{o3} popularized the idea of \emph{thinking with images}, where MLLMs actively operate on images via tools. Most follow-ups focus on the crop/zoom tool~\citep{zheng2025deepeyes,lai2025mini,su2025pixel,fan2025grit}, in part because grounding-rich pretraining eases integration~\citep{bai2025qwen2}. However, cropping is often unnecessary or the benefits are hard to observe for many tasks because similar performance can be achieved without it. Meanwhile, broader tool suites (\textit{e.g.,} line drawing, OCR, segmentation)~\citep{su2025openthinkimg,zhang2025thyme,liu2025visual} have been investigated, but their effectiveness has not been sufficiently validated because they are not evaluated on tasks that truly \emph{require} tools.

\textbf{Tool Integration.} Equipping large models with tools~\citep{comanici2025gemini,o3,gpt5} mitigates intrinsic limitations by enabling external capabilities such as search, code execution, and generative models. For instance, LLM-I~\citep{guo2025llm} leverages tools such as search, code, and diffusion models, enabling a pure language model to produce multimodal content, and recent work integrates web search to support fact-grounded reasoning~\citep{jin2025search,li2025search}. DeepResearch~\citep{team2025tongyi,oaidr} is a more advanced tool-integrated reasoning paradigm, which requires multi-turn search, data collection, and reasoning and achieves state-of-the-art performance on various difficult benchmarks. These directions highlight the promise of tool-augmented agents, yet most image-centric systems still rely on narrow, pre-registered tool sets with hand-specified interfaces.

\textbf{MLLM Reasoning.} Reinforcement learning has become central to strengthening reasoning in LLMs, from PPO~\citep{schulman2017proximal} to more recent GRPO~\citep{shao2024deepseekmath}, DAPO~\citep{yu2025dapo}, and GSPO~\citep{zheng2025group}. For MLLMs, reasoning spans both text and vision, such as chart reasoning~\citep{wang2025omni,masry2025chartqapro}; recent work~\citep{guo2025observe} emphasizes careful visual inspection prior to reasoning via an explicit observe tag. Similarly, a series of works~\citep{hong2025apo, gao2025soft} have explored enhancing reasoning in MLLMs through advanced policy optimization. Recent efforts are transitioning from thinking \emph{about} images to thinking \emph{with} images~\citep{o3,su2025openthinkimg,zhang2025thyme}, but robust, multi-turn, multi-tool composition and strong generalization to unseen tools remain open challenges.

\begin{figure}
    \centering
    \includegraphics[width=0.95\linewidth]{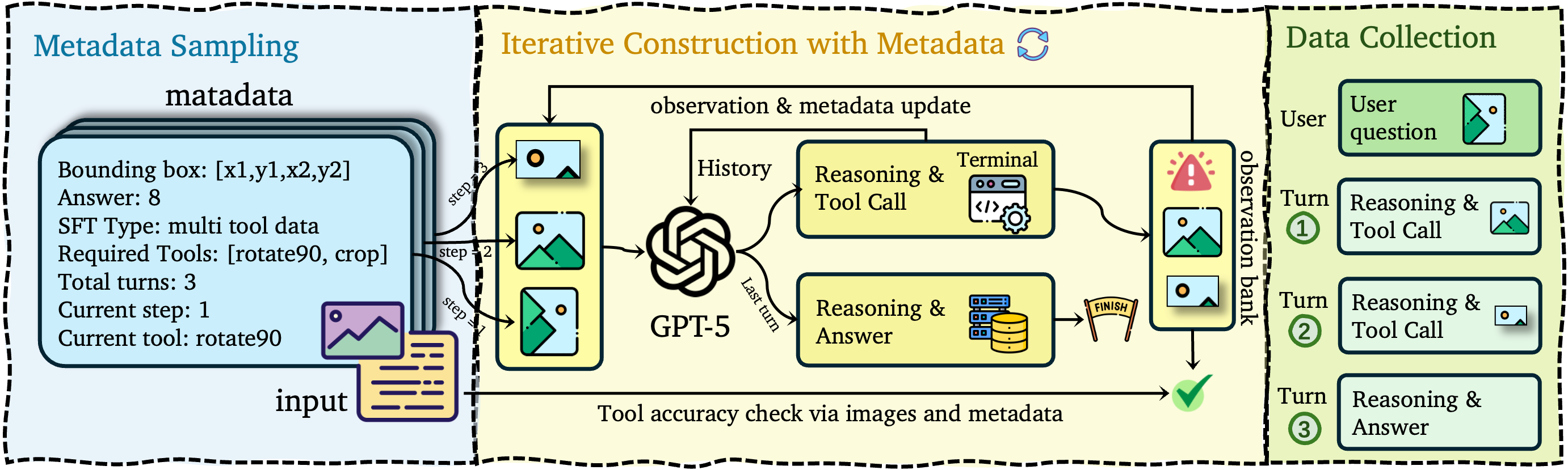}
    \caption{Pipeline for cold-start SFT data construction.}
    \label{fig:sftdata}
\end{figure}

\section{Methodology}

\subsection{Overview}
We train an agent to use tools robustly and efficiently through a two-stage process. We begin with a cold-start phase, where Supervised Fine-Tuning (SFT) teaches the model the fundamental syntax and patterns of tool use on a diverse, high-quality dataset that explicitly covers challenging scenarios such as multi-tool sequences, error handling, and coarse-to-fine localization. Following SFT, we transition to a Reinforcement Learning (RL) phase to move beyond pattern imitation and teach the model to strategize. In this stage, the agent learns to make deliberate, efficient, and robust decisions via a dense, multi-component reward function designed to encourage the use of necessary tools, facilitate the discovery of beneficial strategies, and penalize reward-hacking behaviors, ensuring the final model is effective and deliberate in its tool application. We next detail the cold-start data and training, followed by the RL setting, including data construction and reward design.

\subsection{Cold Start}

\subsubsection{Dataset Construction}
\label{sec:dataset}
The overall pipeline for cold-start SFT data construction is shown in Figure~\ref{fig:sftdata}.

We build an SFT corpus that couples tool-use trajectories with final answers by combining diverse sources, explicit metadata, and programmatic verification. As noted in Figure~\ref{fig:diagnostic}, even state-of-the-art models struggle to reliably determine which tool is needed, so we guide data generation with structured metadata and automatic checks.

\textbf{Data sources across domains.} To increase diversity, we aggregate examples from multiple domains: handwriting datasets~\citep{marti2002iam,liu2011casia}, in-the-wild OCR/VQA datasets for everyday scenes~\citep{long2022towards}, table and chart understanding datasets~\citep{masry2023unichart}, and math reasoning datasets~\citep{qiao2025we}.

\textbf{Task types and metadata.} For each sample, we create metadata containing the ground-truth answer and a target \emph{type} from five categories: \texttt{single-tool}, \texttt{multi-tool}, \texttt{multi-crop}, \texttt{error-handling}, and \texttt{no-tool}. We sample types according to preset proportions and, conditioned on the type, uniformly sample candidate tools. For crop-type data, we select text regions annotated in the source dataset whose areas are at most 0.01\% of the image, ensuring that cropping is necessary for successful recognition and reasoning. For the multi-crop type, we simulate coarse-to-fine zoom-in by enforcing monotonically shrinking, spatially contiguous crop windows across steps to strengthen spatial awareness and localization. For the error-handling type, we intentionally surface failures, such as using an incorrect tool or triggering code and runtime errors (\textit{e.g.,} invalid arguments, missing imports), and require the model to read error logs, revise its code, and retry—potentially switching to the correct tool—with limited retries before fallback.

\textbf{Metadata-conditioned image transformations.} After sampling the metadata, we transform the canonical (unaltered) image according to the selected tool to construct the model's initial observation. For example, if the required tool is \texttt{rotate-180}, we rotate the original image by 180 degrees and use this transformed image as GPT-5's initial input. This setup makes the subsequent tool invocation necessary to recover the canonical view.

\textbf{SFT generation with GPT-5.} Given the question, the (possibly transformed) image, and the metadata, GPT-5 produces a step-by-step reasoning trace and an \emph{action} at each turn. Actions may include calling a tool or emitting an intermediate or final answer. When an action calls a tool, we execute it in a controlled runtime to obtain an updated image. We then compare the output against the canonical, untransformed reference. If they agree, the tool call is marked correct. Otherwise, we either discard the trajectory or attempt targeted correction. For the error-handling type, we feed error logs back to the model to trigger code revision before any discard.

\textbf{Iterative multi-turn construction.} After each step, we update the metadata (step count, remaining/required tools) and resubmit the updated context to GPT-5, repeating until the per-type turn budget is reached or a \texttt{stop} action is produced. Following this pipeline, we construct approximately 5,000 high-quality SFT examples consisting of aligned reasoning, action sequences, and answers.

\begin{figure}
    \centering
    \includegraphics[width=\linewidth]{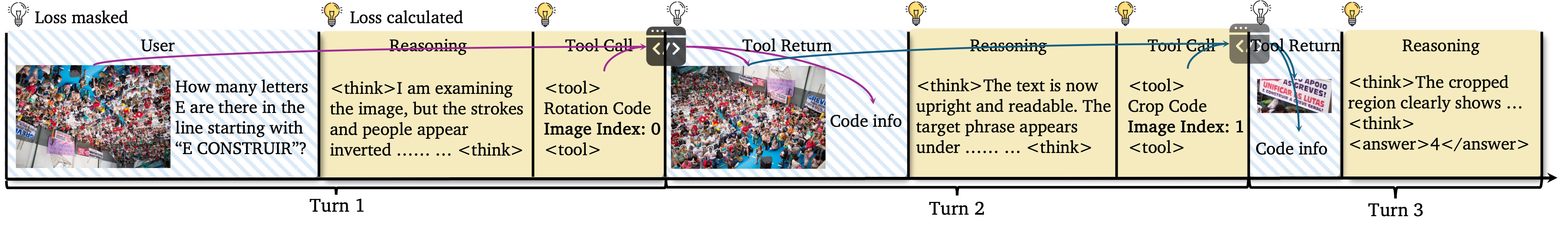}
    \caption{Rollout and inference process, and token masking used during SFT/RL.}
    \label{fig:rollout}
\end{figure}

\subsubsection{Training}

We train the model with an SFT objective over our multi-turn trajectories. As shown in Figure~\ref{fig:rollout}, each training example is formatted as an interleaved dialogue with alternating user prompts, assistant reasoning/tool-call outputs, and recorded tool returns. During optimization, we \emph{mask out} user tokens and tool-return tokens; only assistant tokens corresponding to chain-of-thought reasoning and tool-call specifications contribute to the loss. Concretely, we apply teacher-forced next-token prediction with a token-wise mask \(m_t \in \{0,1\}\) that is 1 for assistant reasoning/tool-call tokens and 0 otherwise.

\begin{equation}
    \label{eq:sft}
    \mathcal{L}_{\mathrm{SFT}}(\theta)
    = - \sum_{t=1}^{T} m_t \; \log p_{\theta}\!\left( y_t \,\middle|\, \mathbf{x}, \mathbf{y}_{<t} \right),
\end{equation}
where \(\mathbf{x}\) denotes the dialogue context (including masked user inputs and tool returns), \(\mathbf{y}_{<t}\) are preceding assistant tokens, and \(m_t\) selects only assistant reasoning/tool-call tokens.

We do not execute tools online during SFT. All tool outputs used as subsequent context are the cached results from dataset construction. Training therefore reduces to standard causal LM training with a masked loss over the assistant side across multiple turns (including the final answer), preserving the multi-turn structure while avoiding runtime variance from external tool execution.

\subsection{Reinforcement Learning}
To further enhance reasoning and generalization, we conduct RL training after the cold-start stage.

\subsubsection{Dataset}
Beyond the SFT sources, we augment RL training data with additional reasoning-heavy and perception samples~\citep{mathew2021docvqa, wang2025sota, singh2019towards} to increase coverage and diversity of reasoning patterns. We then perform difficulty filtering to focus optimization on informative instances. For each candidate item, we sample multiple rollouts and remove degenerate cases where trajectories are uniformly trivial (all-correct) or uninformative (all-incorrect). This concentrates training on items with signal for policy improvement. Furthermore, each item is annotated with a \emph{must-use} tool field drawn from \{\texttt{rotate90}, \texttt{rotate180}, \texttt{rotate270}, \texttt{flip-horizontal}, \texttt{flip-vertical}, \texttt{crop}\}. For crop-required items, we additionally attach the bounding box of the target region. Items that require no tools set this field to \texttt{None}. These constraints encourage policies to learn when and how to invoke the right tool. Following this pipeline, we construct approximately 40,000 RL training items.

\subsubsection{Reward Function}

Training LLMs to use tools reliably is challenging in practice: training collapse and abnormal tool-call rates frequently occur. We therefore adopt a dense, rich, and multi-component reward to stabilize and guide RL. Let a trajectory be \(\tau = (s_1,a_1,\dots,s_T,a_T)\), where \(s_t\) is the environment state (dialogue context, current image, and any tool returns) and \(a_t\) is the agent action (reasoning/tool-call program or textual answer) at turn $t$. The total reward decomposes as
\begin{equation}
    \label{eq:rl_total}
    R_{\text{total}}(\tau)
    = R_{\text{outcome}}(\tau)
    + \beta_1 \sum_{t=1}^{T} \ R_{\text{strategy}}(a_t) - \beta_2 P_{\text{cost}}(\tau).
\end{equation}
where \(R_{\text{outcome}}(\tau)\) is the outcome reward consisting of the terminal accuracy signal and the formatting reward, \(R_{\text{strategy}}(a_t)\) is the process reward for strategy supervision at turn $t$, \(P_{\text{cost}}(\tau)\) is the constraint penalty for trajectory efficiency and avoiding reward hacking, and $\beta_1, \beta_2$ control the trade-off between these terms.

\textbf{Outcome Reward.} We use a terminal accuracy signal \(r_{\text{acc}}\in\{0, +1\}\) assessed on the final answer, and a format reward \(r_{\text{fmt}}\in\{0, +1\}\) for correct reasoning formatting (\texttt{<think>} and \texttt{<answer>} tags).

\textbf{Strategy Shaping.} Pure outcome rewards are sparse and often lead to brittle behavior (\textit{e.g.,} abnormal tool-call rates). We therefore introduce process-level signals with two parts: (i) a \emph{must-use} tool set that encodes task prerequisites, and (ii) \emph{suggested} tools discovered on-the-fly that empirically improve success.

\emph{(a) Must-use tools \(S_{\text{req}}\).} Our RL metadata specifies a set of tools, \(S_{\text{req}}\), that are prerequisites for solving the task robustly. For example, an image with incorrect orientation must be rotated first, regardless of the question. Let \(N=|S_{\text{req}}|\) be the number of required tools. We assign a total reward budget of \(1/N\) to each required tool.
For categorical tools like ``rotate'' or ``flip'', this reward is given as a one-time bonus of \(1/N\) when the tool is correctly used for the first time. For ``crop'', which has a continuous quality measure, the reward is proportional to the Intersection-over-Union (IoU)---a measure of overlap between the predicted and target bounding boxes. To encourage refinement over multiple steps, we only reward the \emph{improvement} in IoU over the best attempt so far. Additionally, we introduce a bonus for perfectly matching the required tool-use trajectory. This reward is granted only if the agent executes the complete sequence of tools specified in \(S_{\text{req}}\) correctly and in the prescribed order, without any redundant or incorrect steps. This mechanism serves two purposes. First, for single-tool scenarios, it encourages precision by distinguishing between an immediate, correct tool call and a solution reached through trial-and-error. Second, for multi-tool tasks, it incentivizes the agent to learn and follow the optimal solution path, ensuring it explores and masters the entire sequence of required actions.

\emph{(b) Suggested tools bonus.} Not all useful tools can be predefined. In addition to the must-use tools in $S_{\text{req}}$, we consider optional tools that are not required but can empirically improve success. This is where the power of our code-as-a-tool framework becomes critical, as it liberates the agent from a fixed tool registry and allows it to invoke any function or library dynamically. For example, applying contrast enhancement on a blurry image might be beneficial but is not a universal prerequisite. We reward the discovery of such emergent, helpful tools through a rollout comparison mechanism.
For a given problem, we collect \(K\) trajectories (with \(K=8\) in our experiments) from the current policy and partition them into those that use an optional tool, $G_\text{tool}$, and those that do not, $G_\text{notool}$. If the tool-using group shows a higher accuracy rate and the no-tool group achieves at most one success out of \(K\) rollouts, it suggests that the additional tools will improve the success rate of solving this problem. We then compute this empirical performance gain as an ``inferred tool necessity reward'' for this problem:
\begin{equation}
    r_{\text{nec}} = \max\!\Big(0,~ \frac{\sum_{i \in G_\text{tool}} r^i_{acc}}{|G_\text{tool}|} - \frac{\sum_{i \in G_\text{notool}} r^i_{acc}}{|G_\text{notool}|} \Big),
\end{equation}
where $|G|$ is the number of trajectories in group $G$. This reward \(r_{\text{nec}}\) is then added as a bonus to all successful trajectories that use the beneficial optional tool, encouraging the agent to explore and adopt various emergent tools that demonstrably improve performance. In addition, to further reward the use of extra tools, we introduce a per-trajectory bonus that is granted whenever the agent both invokes at least one optional tool and produces a correct final answer.

\textbf{Constraint Penalties.} While dense rewards are effective at guiding the agent, they can also be exploited. We observe emergent ``reward hacking'' behaviors during training, where the agent would learn to maximize rewards in ways that misalign with our goal of efficient and intelligent problem-solving. For instance, an agent might continue to crop an image to fractionally improve its IoU score long after the correct answer has been determined, or attempt to rotate an already correctly-oriented image to seek a reward. These actions lead to inefficient, low-quality trajectories. To counteract this, we introduce three targeted penalties that act as guardrails:

\begin{itemize}
    \item \textbf{Turn Limit Penalty:} The primary motivation for this penalty is to prevent reward hacking and improve efficiency. We observe that the agent could learn to exhaustively call all possible required tools (\textit{e.g.,} trying rotate90, rotate180, and rotate270 sequentially) simply to maximize its \(R_{\text{strategy}}\) score, even if the first action already solves the orientation problem. To curb this behavior and promote deliberate, accurate solutions, we penalize trajectories that use more tool-call turns than necessary. A problem with \(|S_{\text{req}}|\) required tools is given a buffer of one extra turn for error handling and exploration; any tool-call turn exceeding \(|S_{\text{req}}|+1\) incurs a penalty.
    
    \item \textbf{Poor Reasoning Penalty:} An agent must arrive at the right answer for the right reasons. If a trajectory yields the correct final answer but relies on a ``crop'' action with a negligible IoU (\(< 0.1\)) with the true region of interest, it indicates that the agent's reasoning was not grounded in the correct visual evidence. We penalize such low-quality trajectories to enforce a stronger link between visual evidence and the final conclusion.
    
    \item \textbf{Inappropriate Tool Use Penalty:} A core aspect of tool use is knowing when \textit{not} to act. For samples where no tools are required (\(S_{\text{req}} = \text{None}\)), any attempt to use an orientation-adjusting tool is penalized. Applying such transformations to a normal image actively increases the task's difficulty and demonstrates a flawed, low-quality strategy.
\end{itemize}

Finally, the constraint penalty $P_{\text{cost}}(\tau)$ consists of the sum of the above three penalties where each penalty $p\in \{0, +1\}$.

\begin{table}[]
\centering
\caption{Main results on OCRBench (Perception) and ChartQAPro (Reasoning) under five types of transformations.}
\label{tab:main_results1}
\resizebox{0.95\textwidth}{!}{%
\begin{tabular}{@{}l|ccccccc|ccccccc@{}}
\toprule
\multicolumn{1}{c}{} & \multicolumn{7}{c}{OCRBench} & \multicolumn{7}{c}{ChartQAPro} \\ \cmidrule(l){2-15} 
\multicolumn{1}{c}{\multirow{-2}{*}{Model}} & Source & Rot90 & Rot180 & Rot270 & Hori & Verti & \cellcolor[HTML]{CBCEFB}Avg & Source & Rot90 & Rot180 & Rot270 & Hori & Verti & \cellcolor[HTML]{CBCEFB}Avg \\ \midrule
GPT-4o & 83.1 & 61.9 & 46.6 & 63.7 & 48.1 & 12.9 & \cellcolor[HTML]{CBCEFB}52.7 & 50.3 & 40.9 & 33.8 & 41.3 & 33.1 & 25.3 & \cellcolor[HTML]{CBCEFB}37.4 \\
Gemini2.5-Pro & 87.1 & 68.2 & 67.9 & 71.5 & 39.4 & 41.3 & \cellcolor[HTML]{CBCEFB}62.6 & 66.8 & 63.8 & 59.7 & 61.6 & 45.6 & 58.2 & \cellcolor[HTML]{CBCEFB}59.3 \\
Qwen2.5-VL-32B & 85.0 & 64.5 & 45.6 & 64.8 & 25.5 & 8.7 & \cellcolor[HTML]{CBCEFB}49.0 & 39.5 & 29.7 & 26.0 & 30.7 & 22.6 & 17.6 & \cellcolor[HTML]{CBCEFB}27.7 \\
Qwen2.5-VL-72B & 88.5 & 72.6 & 58.3 & 73.5 & 35.1 & 18.2 & \cellcolor[HTML]{CBCEFB}57.7 & 38.2 & 29.5 & 27.7 & 30.3 & 23.8 & 20.3 & \cellcolor[HTML]{CBCEFB}28.3 \\
Qwen3-VL-30B-Thinking & 87.6 & 72.1 & 60.5 & 66.6 & 49.3 & 31.8 & \cellcolor[HTML]{CBCEFB}61.3 & 50.2 & 38.7 & 36.5 & 38.2 & 31.3 & 29.0 & \cellcolor[HTML]{CBCEFB}37.3 \\
Qwen3-VL-235B-Thinking & 88.8 & 76.4 & 71.0 & 74.3 & 45.9 & 23.8 & \cellcolor[HTML]{CBCEFB}63.4 & 56.9 & 45.0 & 43.4 & 46.9 & 35.1 & 26.0 & \cellcolor[HTML]{CBCEFB}42.2 \\
InternVL3.5-30B & 88.7 & 45.8 & 7.2 & 45.8 & 5.9 & 6.2 & \cellcolor[HTML]{CBCEFB}33.3 & 37.2 & 27.2 & 18.0 & 27.9 & 17.5 & 18.0 & \cellcolor[HTML]{CBCEFB}24.3 \\
InternVL3.5-241B & 92.3 & 58.2 & 32.4 & 57.6 & 25.6 & 9.4 & \cellcolor[HTML]{CBCEFB}45.9 & 45.9 & 31.5 & 24.4 & 34.7 & 22.5 & 21.3 & \cellcolor[HTML]{CBCEFB}30.0 \\
Thyme & 86.3 & 67.0 & 51.9 & 67.8 & 27.2 & 13.3 & \cellcolor[HTML]{CBCEFB}52.3 & 30.3 & 23.8 & 20.4 & 22.9 & 17.8 & 14.4 & \cellcolor[HTML]{CBCEFB}21.6 \\\midrule
Qwen2.5-VL-7B & 86.4 & 70.2 & 58.0 & 71.7 & 32.4 & 17.0 & \cellcolor[HTML]{CBCEFB}56.0 & 37.3 & 23.4 & 22.2 & 23.7 & 19.5 & 20.1 & \cellcolor[HTML]{CBCEFB}24.4 \\
\rowcolor[HTML]{FFCCC9} 
CodeVision-7B & 87.2 & 72.3 & 73.1 & 75.2 & 65.1 & 67.4 & 73.4 & 39.1 & 30.8 & 29.8 & 31.4 & 30.1 & 29.0 & 31.7  \\
Qwen3-VL-8B-Thinking & 82.4 & 64.8 & 57.6 & 58.4 & 35.7 & 14.5 & \cellcolor[HTML]{CBCEFB}52.2 & 47.2 & 32.8 & 30.0 & 31.9 & 21.3 & 13.9 & 29.5\cellcolor[HTML]{CBCEFB} \\
\rowcolor[HTML]{FFCCC9} 
CodeVision-8B & 83.5 & 78.6 & 77.4 & 76.7 & 68.7 & 67.3 & 75.4 & 50.3 & 39.0 & 38.1 & 39.2 & 39.7 & 38.0 & 40.7 \\
Qwen3-VL-32B-Thinking & 86.6 & 67.8 & 64.3 & 66.4 & 35.1 & 13.8 & 55.7\cellcolor[HTML]{CBCEFB} & 52.3 & 39.9 & 38.3 & 42.0 & 26.1 & 18.7 & 36.2\cellcolor[HTML]{CBCEFB} \\
\rowcolor[HTML]{FFCCC9} 
CodeVision-32B & 87.8 & 82.8 & 79.3 & 81.1 & 77.7 & 68.3 & 79.5 & 57.4 & 54.4 & 53.6 & 54.7 & 52.8 & 53.1 & 54.3 \\  \bottomrule
\end{tabular}%
}
\end{table}

\section{Experiments}

\subsection{MVToolBench}\label{sec:mvtoolbench}

To rigorously evaluate a model's ability to compose multiple tools in sequence, we construct the MVToolBench. The construction process is designed to create challenging scenarios that necessitate the combined use of multiple tools. The pipeline consists of three main stages: data filtering, question generation, and multi-tool augmentation.

\textbf{Data Source and Filtering.}
We build our benchmark upon the HierText dataset~\citep{long2022towards}, which provides rich annotations including bounding box coordinates and corresponding text for words, lines, and paragraphs. To ensure that our benchmark effectively tests the model's ability to perceive fine-grained details, we first perform a filtering step based on the relative size of the text annotations. For each image, we calculate the area of every annotated word, line, and paragraph as a percentage of the total image area. We then retain only the annotations that are exceptionally small. For instance, for word-level questions, only words occupying less than 0.01\% of the total image area are selected as candidates. This filtering serves two critical purposes: 1) it guarantees a high level of difficulty, as the target text is not easily visible in the original view, and 2) it enforces a strong dependency on the ``crop'' tool, as zooming in becomes a prerequisite for successful recognition and reasoning.

\textbf{Question and Answer Generation.}
Using the filtered, small-sized text annotations, we programmatically generate a variety of reasoning questions. These questions cover tasks such as text recognition (\textit{e.g.,} ``What does the line beginning with `Busy' say?''), counting (\textit{e.g.,} ``How many times does the letter `a' appear in the paragraph ending with `Television'?''), and question answering that requires information retrieval from specific paragraphs. A crucial design principle during this stage is the deliberate avoidance of any explicit positional cues. We do not use phrases like ``the word on the left'' or provide bounding box coordinates in the prompts. This forces the model to rely entirely on its own grounding capabilities to first locate the relevant region and then perform an accurate crop, significantly increasing the challenge of the benchmark.

\begin{table}[]
    \centering
    \caption{Results on single-tool (V*, HRBench) and multi-tool (MVToolBench) benchmarks.}
    \label{tab:main_results2}
    \resizebox{0.55\textwidth}{!}{%
    \begin{tabular}{@{}l|cccc@{}}
    \toprule
    Model & V* & HRBench4k & HRBench8k & MVToolBench \\ \midrule
    GPT-4o & 67.9 & 65.0 & 60.1 & 8.5 \\
    Gemini2.5-Pro & 83.8 & 86.2 & 85.1 & 32.6 \\
    Qwen2.5-VL-32B & 81.9 & 73.8 & 70.5 & 16.4 \\
    Qwen3-VL-30B-Thinking & 81.2 & 77.8 & 71.3 & 23.7 \\
    Qwen3-VL-235B-Thinking & 85.9 & 84.3 & 76.6 & 30.1 \\
    Thyme & 82.2 & 77.0 & 72.0 & 24.2 \\\midrule
    Qwen2.5-VL-7B & 74.6 & 69.4 & 67.5 & 18.1 \\
    \rowcolor[HTML]{FFCCC9} CodeVision-7B & 83.7 & 75.6 & 72.2 & 60.1 \\
    Qwen3-VL-8B-Thinking & 77.5 & 72.4 & 68.1 & 19.7 \\
    \rowcolor[HTML]{FFCCC9} CodeVision-8B & 82.4 & 77.1 & 73.4 & 62.7 \\ 
    Qwen3-VL-32B-Thinking & 84.8 & 82.1 & 74.8 & 28.6 \\
    \rowcolor[HTML]{FFCCC9} CodeVision-32B & 86.2 & 84.3 & 76.1 & 65.4 \\ 
    \bottomrule
    \end{tabular}%
    }
\end{table}

\textbf{Multi-Tool Augmentation.}
After the question-answer pairs are finalized, we perform a final augmentation step on the images to create a multi-tool requirement. Each image is randomly subjected to one of several transformations: rotation (90, 180, or 270 degrees), horizontal flip, or vertical flip. This step ensures that solving a problem requires not only a ``crop'' operation but also an initial orientation-correction tool. By setting the proportion of each transformation to be equal, we maintain a balanced and unbiased dataset. This three-stage process results in a comprehensive benchmark that effectively evaluates the model's capacity for complex, multi-step tool composition.
\begin{wrapfigure}{r}{6.5cm}
    \centering
    \includegraphics[width=0.38\textwidth]{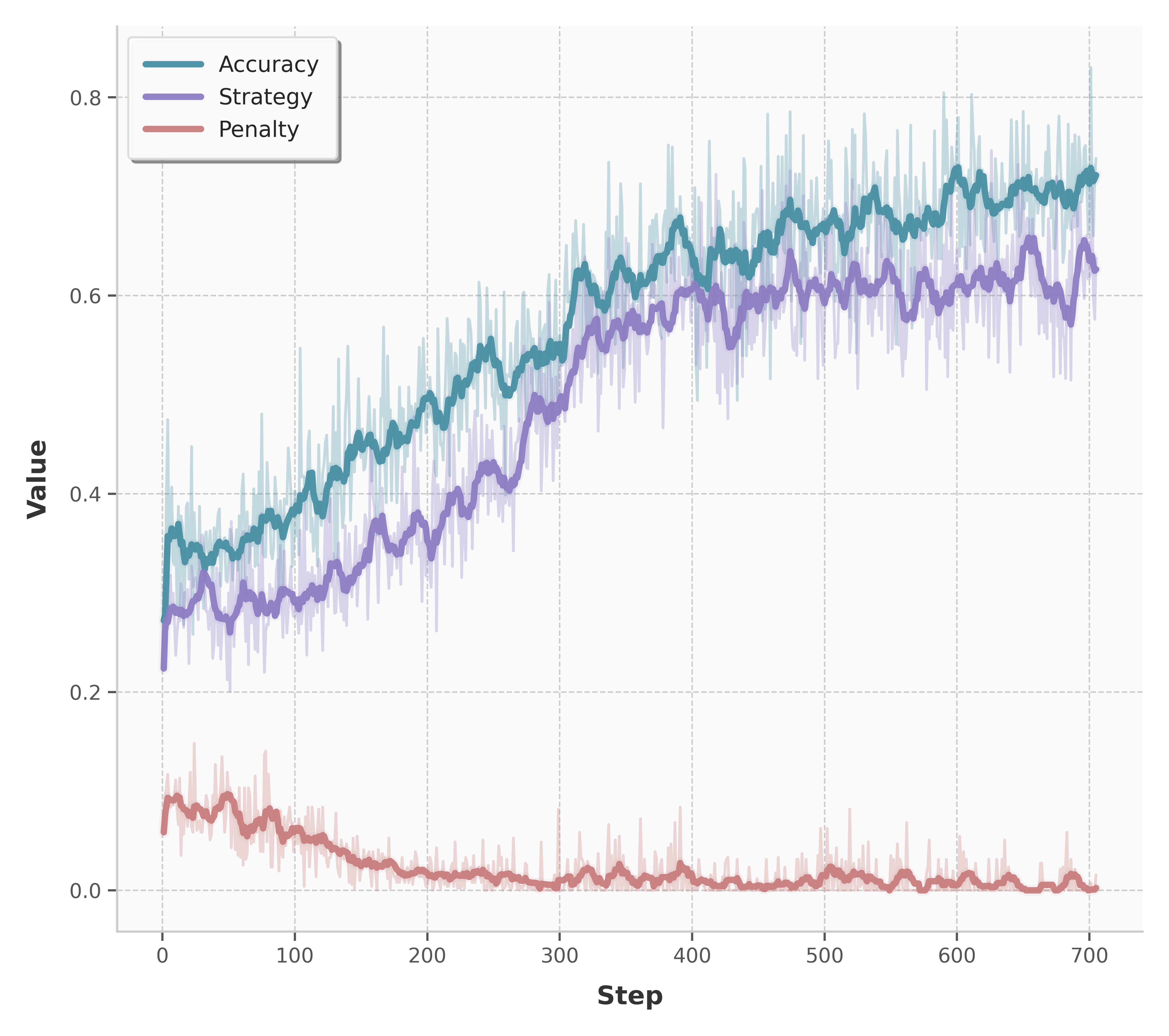}
    \caption{RL training curves for outcome, strategy, and total rewards. The consistent upward trend demonstrates that the agent effectively learns to use tools strategically to solve tasks.}
    \label{fig:mainreward}
    \vskip -0.2in
\end{wrapfigure}
\subsection{Implementation Details}
\textbf{Training Settings: } 
To demonstrate the generalizability of our method, we use three distinct backbone models: Qwen2.5-VL-7B~\citep{bai2025qwen2}, Qwen3-VL-8B-Thinking, and Qwen3-VL-32B-Thinking~\citep{bai2025qwen3}. We employ GRPO~\citep{shao2024deepseekmath} as the base RL algorithm. In the initial SFT stage, we train for 2 epochs with a batch size of 128, a learning rate of 5e-6, a cosine learning rate scheduler, and a warmup ratio of 0.05. Building on the SFT-tuned checkpoint, we then conduct the RL training phase for 2 epochs. For RL, the learning rate is set to 1e-6, with a batch size of 64, and we perform 8 rollouts for each sample. The coefficients for the format reward, strategy reward (must-use and suggested tools), and constraint penalties are set to 0.1, 1.0 (1.0 and 0.2), and 0.5, respectively. The KL divergence coefficient is set to 0.001.

\textbf{Evaluation and Benchmarks: } 
Our evaluation is designed to test tool use across a range of scenarios. For the crop tool, we follow prior work and evaluate on V*~\citep{wu2024v}, HRBench4k, and HRBench8k~\citep{wang2025divide}. To assess the model's ability to handle orientation changes, we create a new benchmark suite by applying five transformations (rotate 90/180/270 degrees, and horizontal/vertical flip) to OCRBench~\citep{liu2024ocrbench} and ChartQAPro~\citep{masry2025chartqapro}. We choose these two datasets to test distinct capabilities: OCRBench is a text recognition benchmark where high performance on the base version indicates that a model must accurately perceive the text to succeed. In contrast, ChartQAPro is a reasoning-based benchmark that requires understanding the chart's content, a task that does not depend on perfectly recognizing every character. This allows us to evaluate performance in both perception-critical and reasoning-heavy scenarios. Furthermore, to evaluate multi-tool proficiency, we utilize our newly created MVToolBench, the construction of which is detailed in Section~\ref{sec:mvtoolbench}.

\begin{figure}
    \centering
    \includegraphics[width=0.7\linewidth]{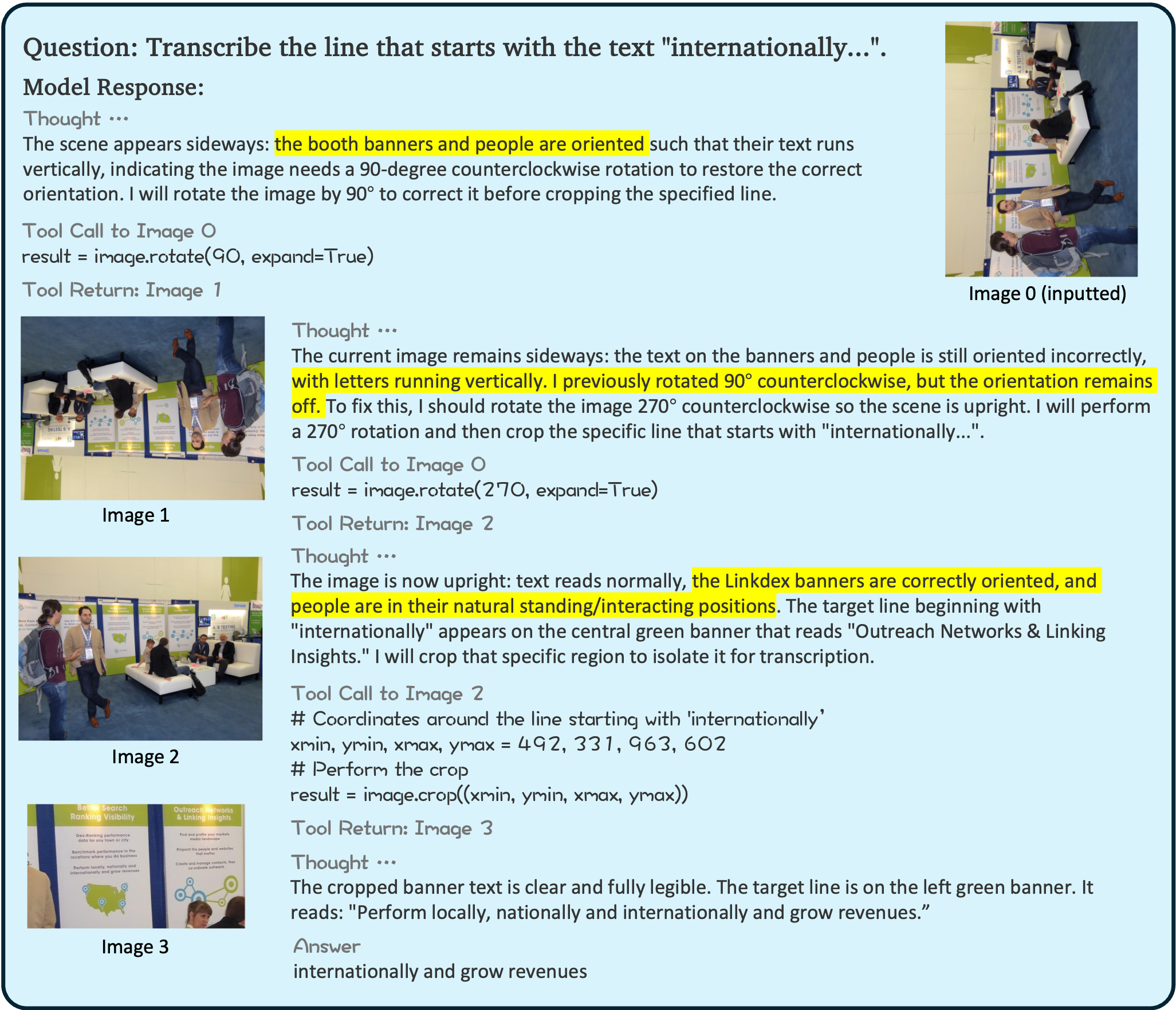}
    \caption{A multi-turn example of error recovery. The model initially calls the wrong tool (\texttt{flip-horizontal}), but after receiving the execution result, it identifies the mistake and corrects it by applying the right tool (\texttt{rotate-90}).}
    \label{fig:exam1}
\end{figure}

\subsection{Main Results}

Table~\ref{tab:main_results1} shows the performance of our CodeVision models against a suite of strong MLLMs on benchmarks with challenging image orientations. On both OCRBench and ChartQAPro, baseline models exhibit a significant performance degradation when faced with rotated or flipped images, confirming the brittleness we identified. In contrast, our CodeVision-7B and CodeVision-8B models consistently and substantially outperform their base models and other strong competitors. For instance, on the transformed OCRBench subset, CodeVision-7B achieves an average score of 73.4, a +17.4 improvement over its base model, demonstrating its robust ability to recognize the correct transformation and apply the corresponding tool to restore the canonical image view for successful reasoning.

\begin{wrapfigure}{r}{6.4cm}
    \centering
    \includegraphics[width=0.35\textwidth]{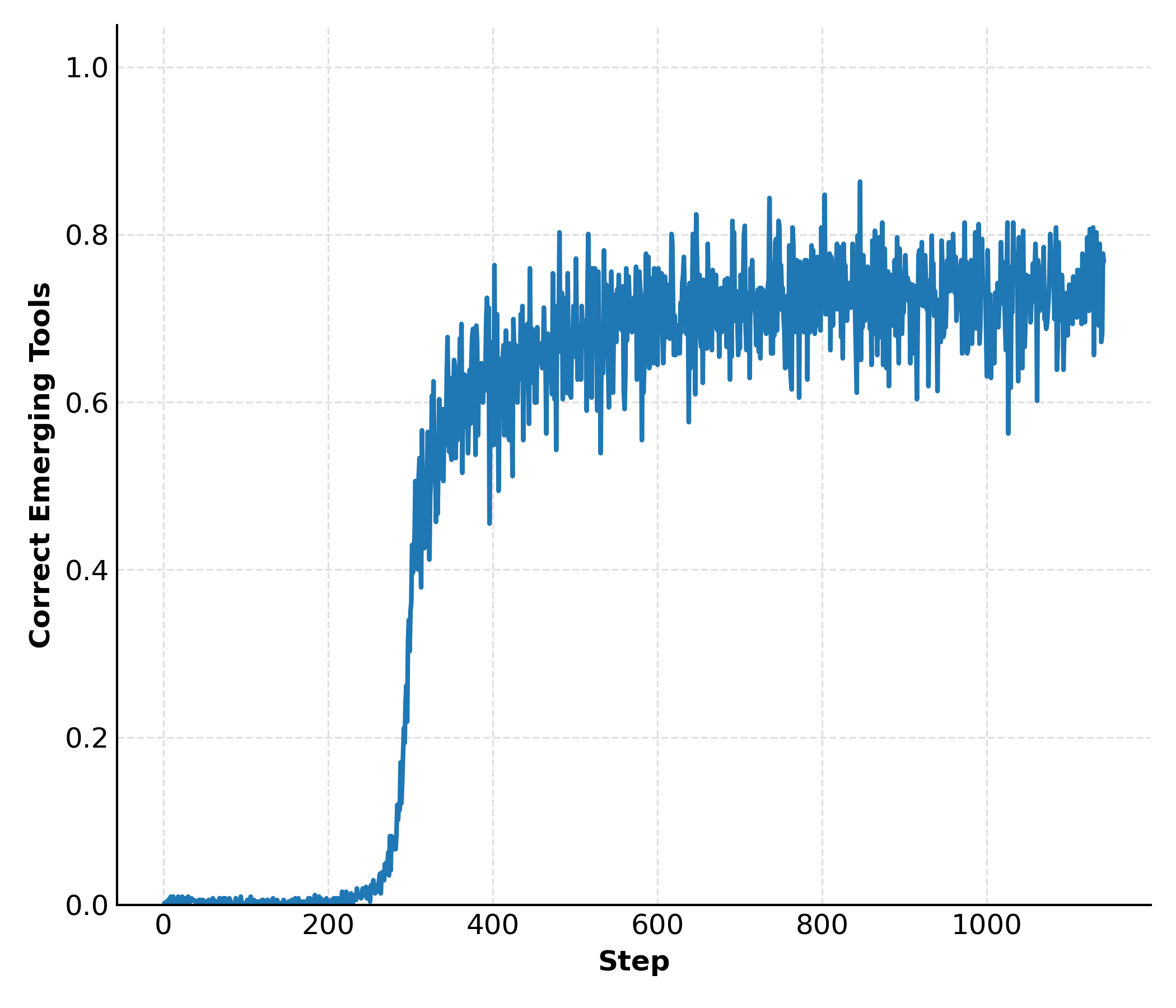}
    \caption{Reward curve for successful emergent tool use during RL training.}
    \label{fig:etoolcurve}
    \vskip -0.1in
\end{wrapfigure}

Table~\ref{tab:main_results2} evaluates performance on established tool-use benchmarks. While our models perform competitively on single-tool tasks requiring cropping (V*, HRBench4k, HRBench8k), they establish a new state-of-the-art on MVToolBench, a benchmark designed for multi-tool reasoning. CodeVision-7B achieves a score of 60.1, nearly doubling the performance of the next-best model, Gemini2.5-Pro (32.6). This highlights our method's superior capability in composing multiple tools to solve complex problems.

The effectiveness of our RL strategy is further illustrated in Figure~\ref{fig:mainreward}, which plots the reward curves during training. The steady increase in all three reward components—outcome, strategy, and the total reward—indicates that the agent is not only learning to achieve the correct final answer but is also mastering the strategic process of tool selection and application as guided by our dense reward function. Additionally, Figure~\ref{fig:etoolcurve} presents the reward curve specifically for successful emergent tool use. This metric tracks instances where the model employs extra tools—beyond the required set—and correctly solves the task. The consistent upward trend signifies that the model is actively discovering and utilizing beneficial tools to enhance its reasoning and success rate, validating the capability of our framework to foster emergent behaviors.

\begin{wrapfigure}{r}{6.4cm}
    \centering
    \includegraphics[width=0.35\textwidth]{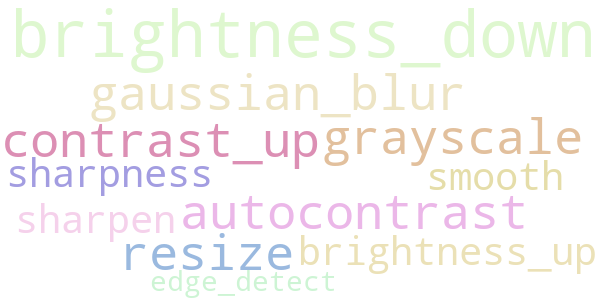}
    \caption{A word cloud of emergent tools discovered during RL training.}
    \label{fig:tools}
    \vskip -0.1in
\end{wrapfigure}

\begin{figure*}
    \centering
    \includegraphics[width=0.7\linewidth]{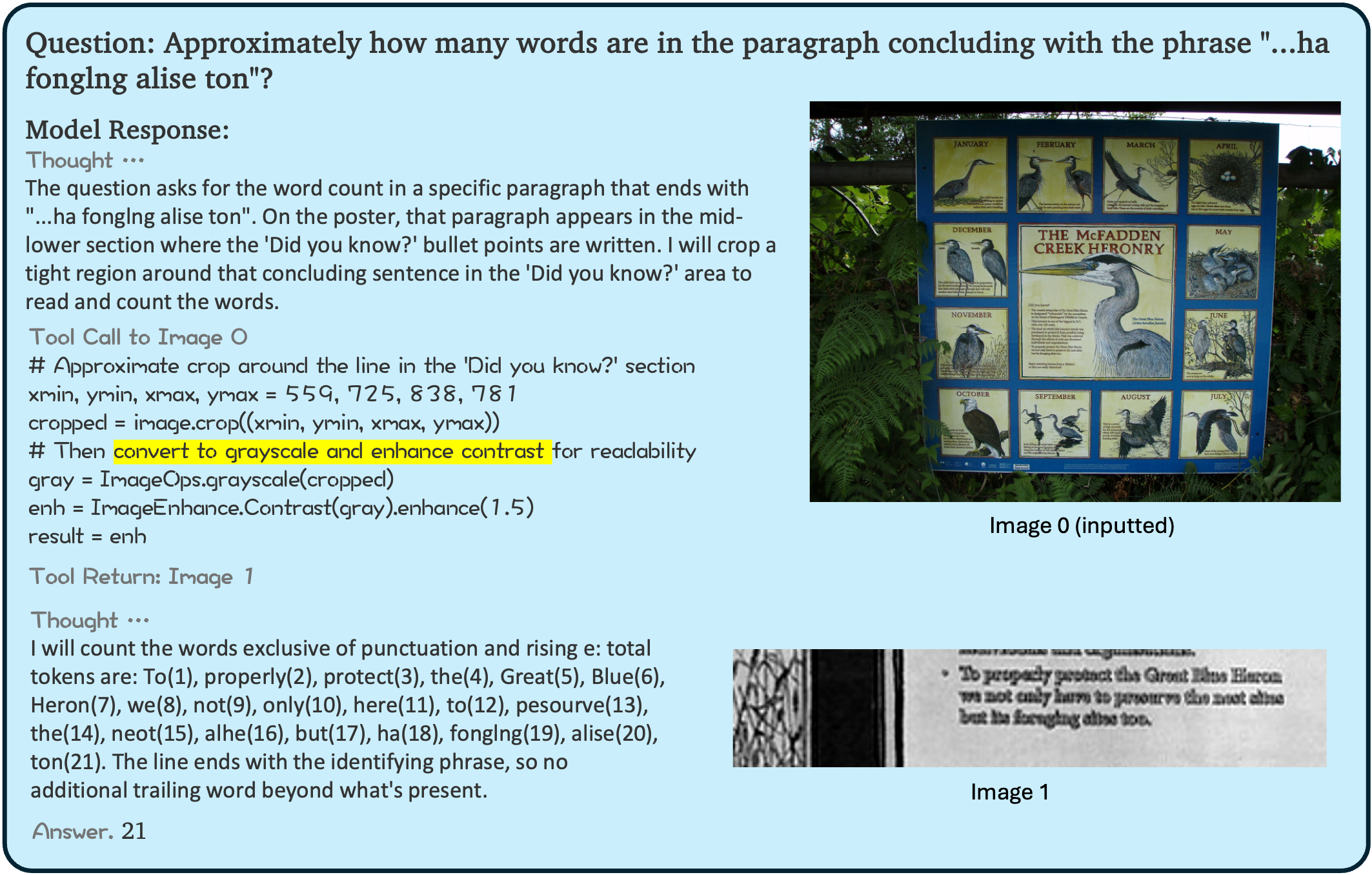}
    \caption{An example of emergent and efficient tool use. The model chains two tools, contrast enhancement and grayscale conversion, within a single turn to fulfill the user's request. These tools did not appear in the RL training set, demonstrating the generalization capability of the code-as-tool framework.}
    \label{fig:exam2}
\end{figure*}

\begin{figure*}[h!]
    \centering
    \includegraphics[width=0.7\linewidth]{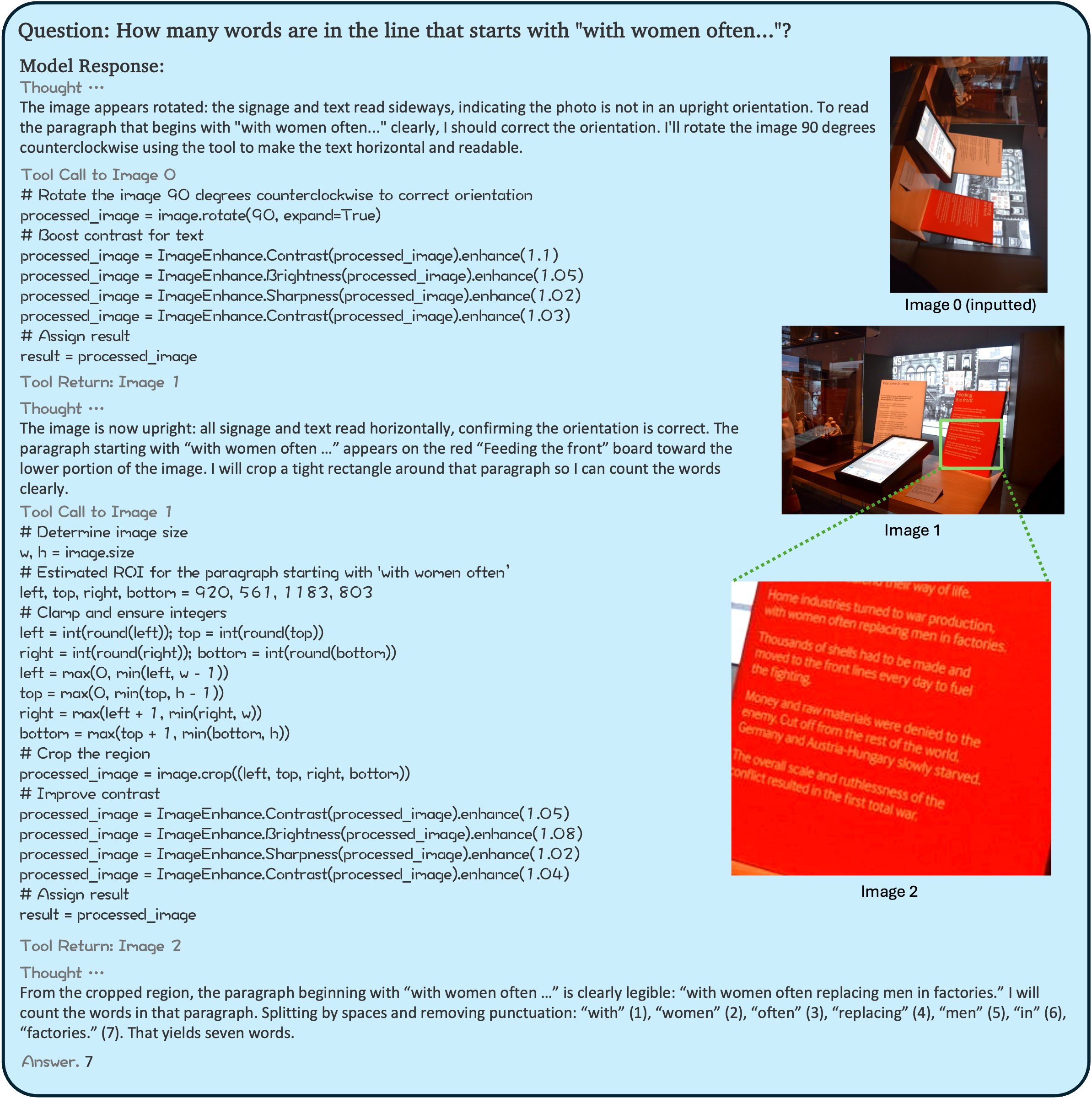}
    \caption{An example of emergent and efficient tool use. The model chains ``brightness up, contrast up, crop, rotate90, sharpness'' tools to solve the user's request.}
    \label{fig:exam7}
\end{figure*}

We provide qualitative examples of our model's advanced capabilities in Figures~\ref{fig:exam1} and~\ref{fig:exam2}. Figure~\ref{fig:exam1} showcases a multi-turn dialogue where the model initially calls an incorrect tool, receives feedback, and robustly recovers by executing the correct tool sequence. This demonstrates sophisticated error-handling and adaptability. Figure~\ref{fig:exam2} highlights two key advantages of our code-as-tool framework: emergent tool use and efficiency. The model chains multiple operations (contrast and grayscale) within a single turn to solve the user's request, even though these specific tools were not included in the RL training data. This ability to generalize and compose novel tool sequences underscores the flexibility of our approach. As shown in Figure~\ref{fig:tools}, our model learns to use a wide variety of other tools not required in our training data, such as adjusting brightness, applying blurs, and detecting edges, demonstrating the near-infinite toolset enabled by the code-as-tool paradigm. Furthermore, Figure~\ref{fig:exam7} presents a striking example where the model spontaneously combines five tools, three of which never appeared in the RL data to solve a request.

\begin{figure*}[h!]
    \centering
    \includegraphics[width=0.7\linewidth]{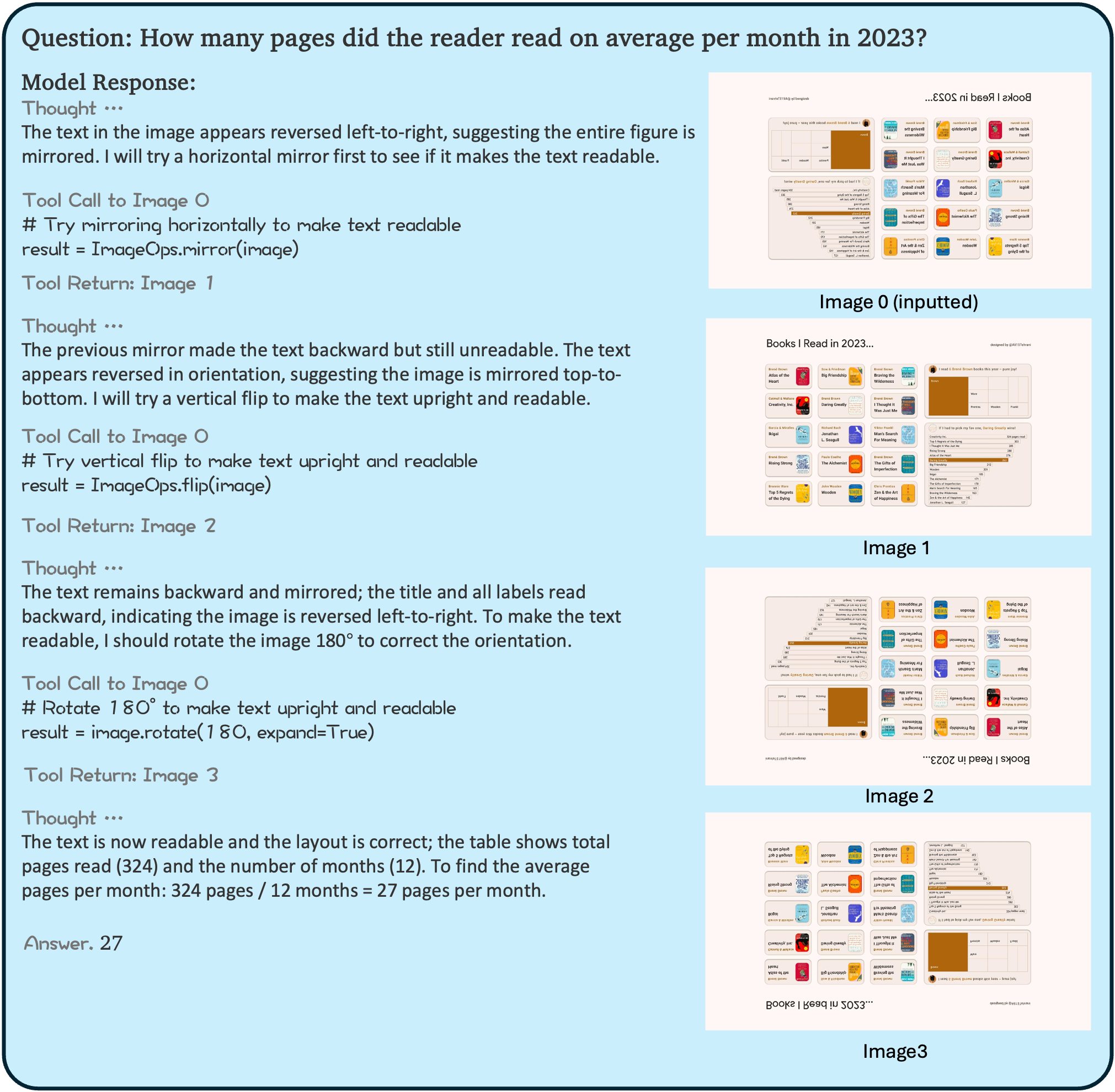}
    \caption{An example of reward hacking from a model trained without constraint penalties. After correctly applying rotate90, the agent continues to call superfluous tools, leading to task failure.}
    \label{fig:exam3}
\end{figure*}

\subsection{Case Study}
To provide a clearer understanding of our model's behavior and the impact of our reward design, we present four case studies.

Figure~\ref{fig:exam3} illustrates a common failure mode known as reward hacking, which we observe when training a model without our proposed constraint penalties. In this example, the agent is tasked with a problem that requires a 90-degree rotation. The agent correctly identifies and applies the rotate90 tool in its first turn, successfully correcting the image's orientation. However, driven solely by the goal of maximizing the strategy reward, it does not stop. Instead, it proceeds to call additional, unnecessary orientation tools in subsequent turns, which corrupts the already-correct image and ultimately leads to failure. This case underscores the critical importance of the penalty functions, which act as guardrails to prevent such inefficient and counterproductive behavior.

\begin{figure*}
    \centering
    \includegraphics[width=0.7\linewidth]{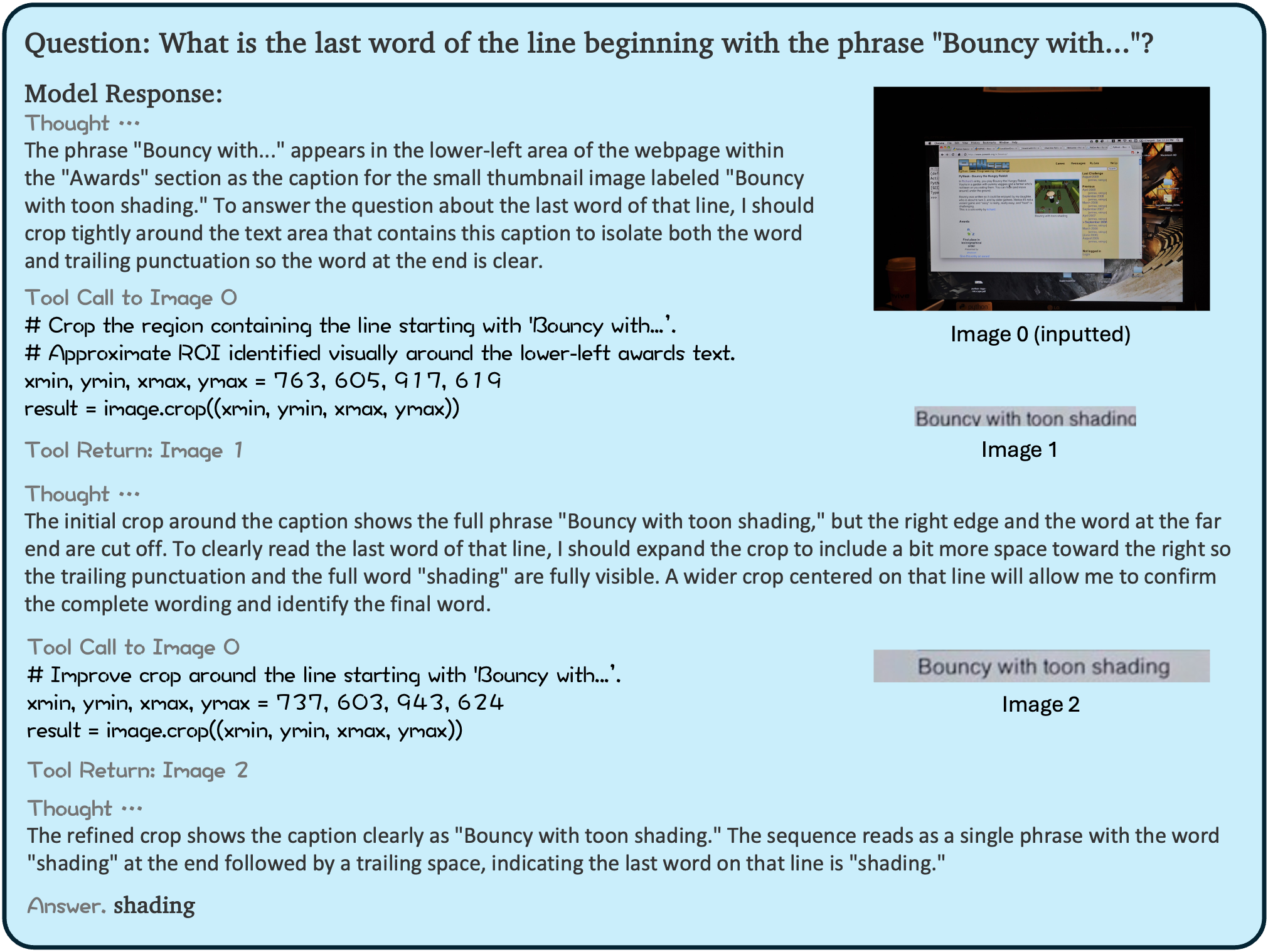}
    \caption{A successful multi-step reasoning case. The model first performs an initial crop, then recognizes its incompleteness and refines the cropped region in a second step to include all necessary information before providing the final answer.}
    \label{fig:exam4}
\end{figure*}

In contrast, Figure~\ref{fig:exam4} showcases the robust, multi-step reasoning capabilities of our fully-trained CodeVision model. The task requires the model to read text from a specific region that is too small to be seen in the initial view. The model first performs a ``crop'' operation to zoom in. Recognizing that its initial crop was incomplete and missed part of the target area, it dynamically adjusts its strategy. In the next turn, it executes a second, more precise ``crop'' that successfully isolates the entire region of interest. With the full context now visible, the model is able to accurately provide the correct answer. This example highlights the model's ability to self-correct and perform fine-grained, iterative adjustments, a key capability for solving complex real-world visual tasks.

\begin{figure*}
    \centering
    \includegraphics[width=0.85\linewidth]{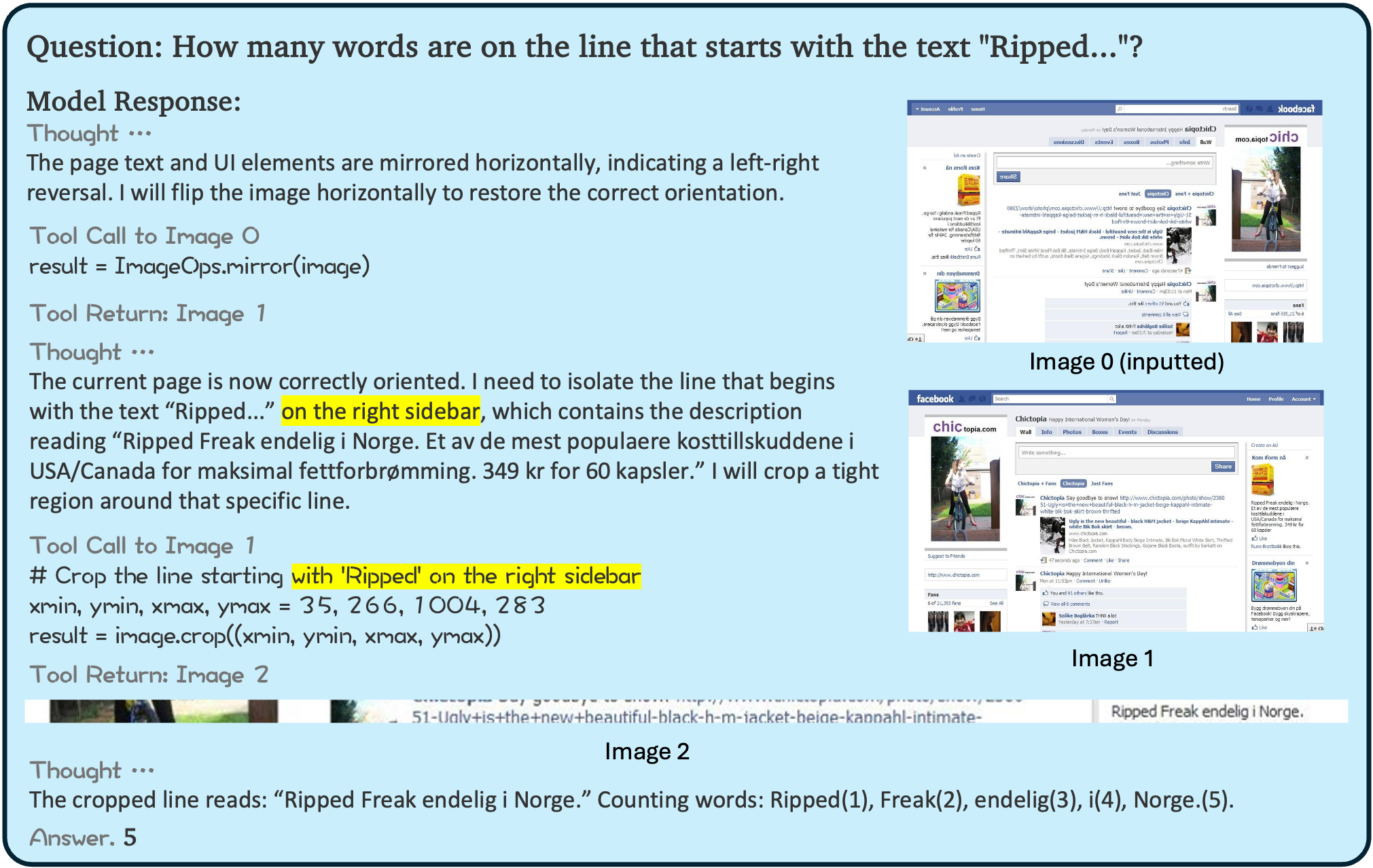}
    \caption{A successful but inefficient localization case. The model corrects the orientation and successfully crops a region containing the target text. However, the crop is a long, narrow strip that includes much irrelevant information, pointing to potential improvements in generating more precise bounding boxes.}
    \label{fig:exam5}
\end{figure*}

Furthermore, Figure~\ref{fig:exam5} presents a nuanced success case that reveals areas for further refinement. The task requires a sequence of orientation correction followed by localizing and reading a very small line of text. The model adeptly handles the first step by correcting the image's orientation. It then correctly reasons about the text's location and performs a crop. Although the cropped region successfully includes the target text, allowing the model to answer correctly, the crop itself is suboptimal—a long, narrow strip containing significant irrelevant context. This example demonstrates that while the model's high-level reasoning is effective, its ability to generate precise, tight bounding boxes could be improved, suggesting it may sometimes adopt an overly cautious safe cropping strategy.

\begin{figure*}
    \centering
    \includegraphics[width=0.85\linewidth]{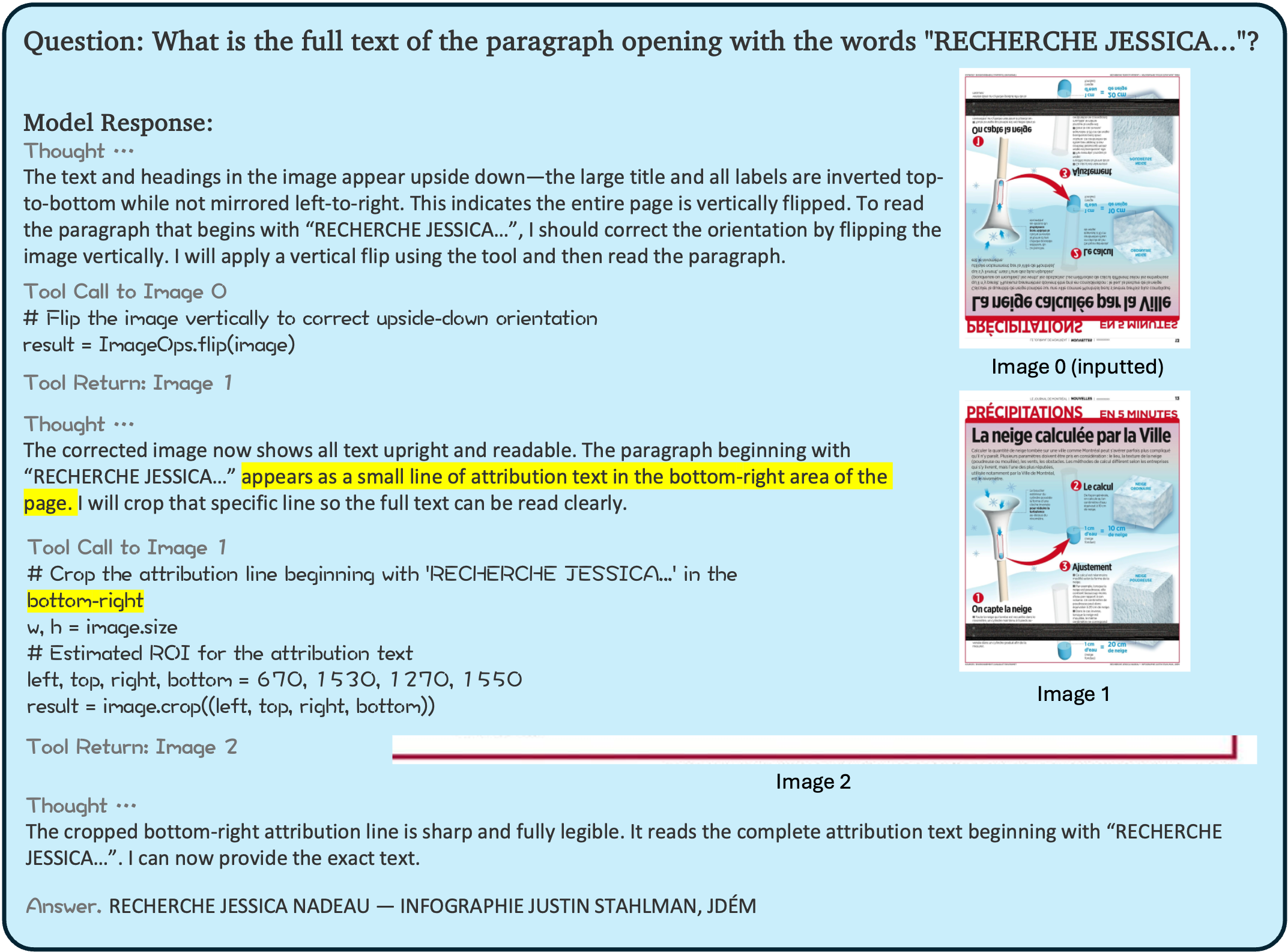}
    \caption{A failure case highlighting the model's limitations in precise localization. The model correctly adjusts the orientation and identifies the general area of interest, but provides slightly inaccurate coordinates for the crop, resulting in a cropped region that just misses the target.}
    \label{fig:exam6}
\end{figure*}

To further probe the model's limitations, we present a clear failure case in Figure~\ref{fig:exam6}. The task requires correcting the image orientation and then cropping a small, specific region to examine details. The model successfully completes the first two stages: it correctly adjusts the orientation and accurately reasons about the approximate location of the region of interest. However, it fails at the final, crucial step of precise localization. The coordinates it generates for the crop are slightly off, causing the cropped area to be adjacent to the target region but missing it entirely. This case clearly demonstrates that while the model excels at high-level reasoning and coarse localization, its fine-grained coordinate prediction remains a challenge and a key area for future improvement.

\subsection{Ablation Study}
To validate the effectiveness of our proposed training methodology and reward components, we conduct a series of ablation studies using the CodeVision-7B model as the baseline.

\begin{figure*}
    \centering
    \begin{subfigure}{0.32\textwidth}
        \centering
        \includegraphics[width=\textwidth]{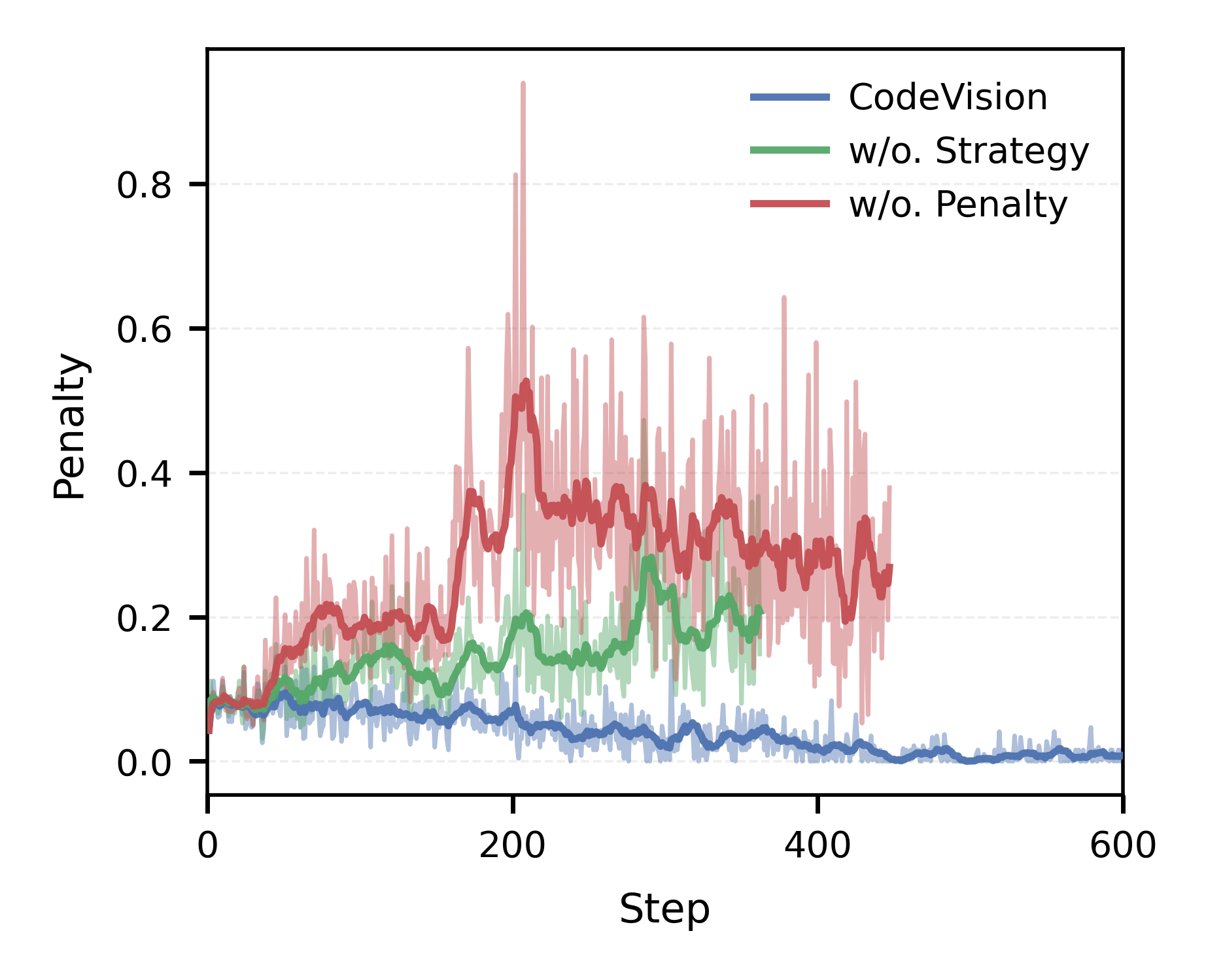}
        \caption{Penalty Term}
        \label{fig:ablation_reward_curves:a}
    \end{subfigure}
    \hfill 
    \begin{subfigure}{0.32\textwidth}
        \centering
        \includegraphics[width=\textwidth]{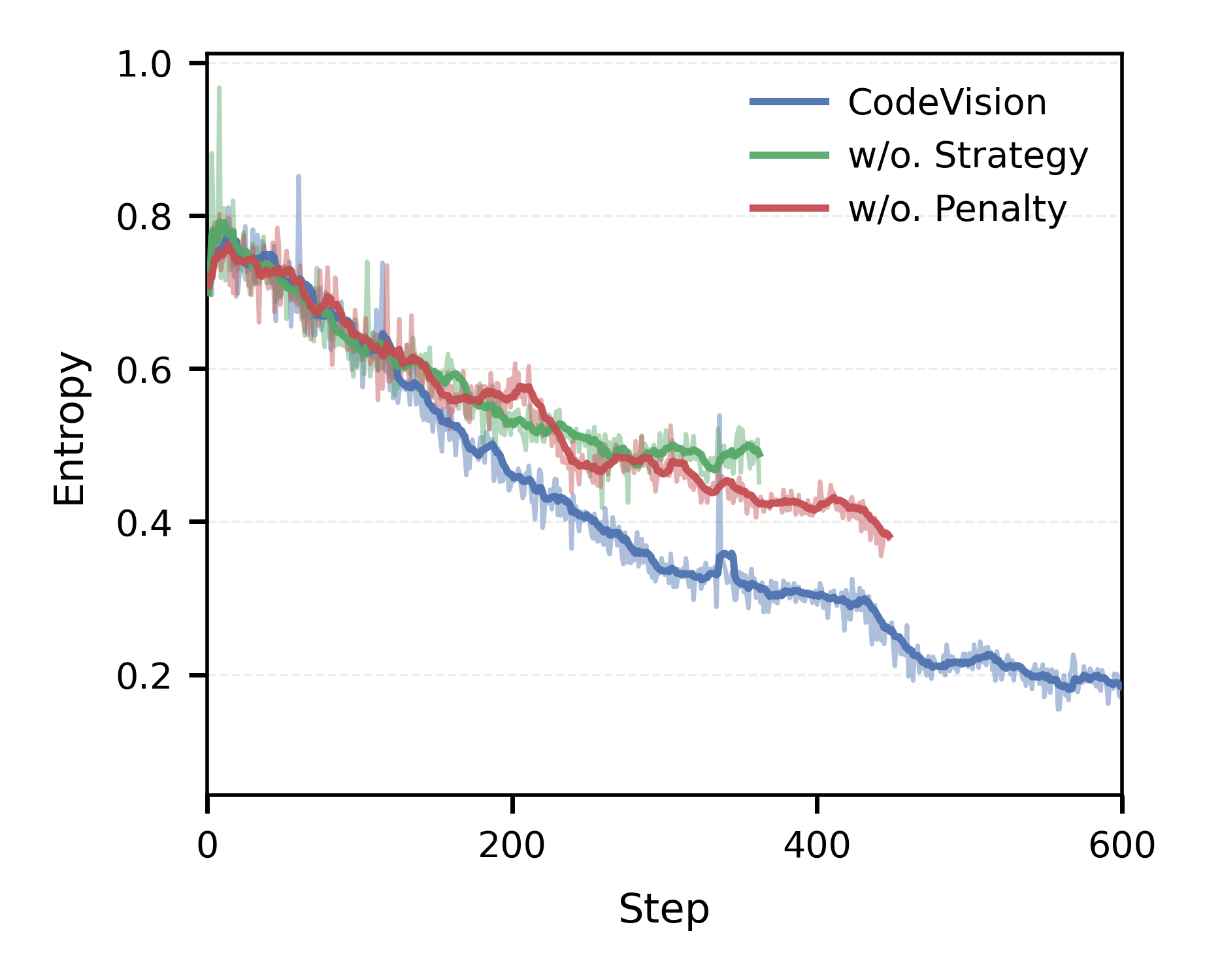}
        \caption{Entropy}
        \label{fig:ablation_reward_curves:b}
    \end{subfigure}
    \hfill
    \begin{subfigure}{0.32\textwidth}
        \centering
        \includegraphics[width=\textwidth]{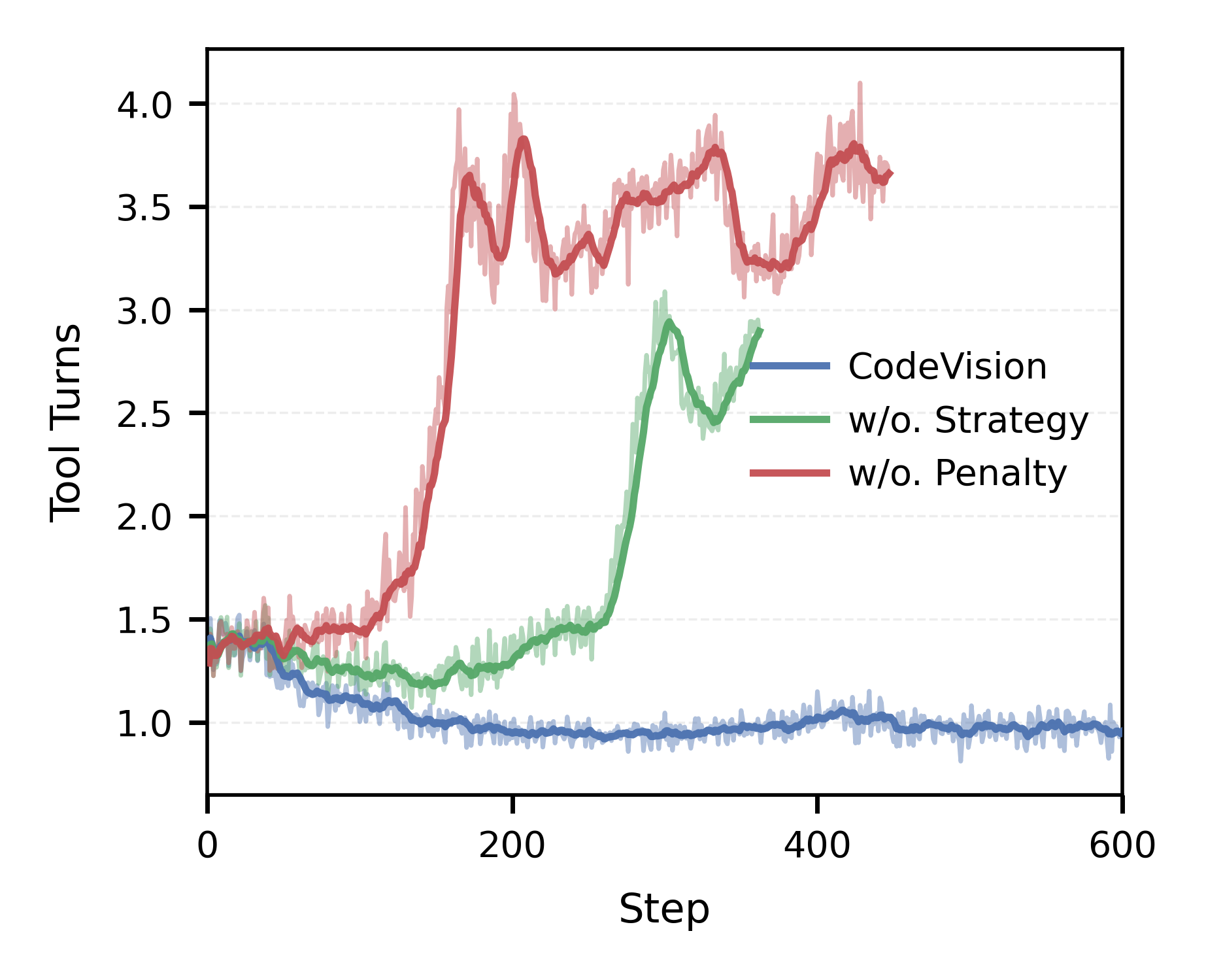}
        \caption{Tool Turns}
        \label{fig:ablation_reward_curves:c}
    \end{subfigure}
    \caption{Training dynamics for our full reward function versus ablated versions. The complete reward function leads to more stable and efficient learning.}
    \label{fig:ablation_reward_curves}
\end{figure*}

\textbf{Effectiveness of dense reward.}
We investigate the impact of our dense reward function by training two variants of our model: one without the strategy-shaping reward (\(R_{\text{strategy}}\)) and another without the constraint penalties (\(P_{\text{cost}}\)). The results are presented in Table~\ref{tab:ablation} and Figure~\ref{fig:ablation_reward_curves}.

As shown in Table~\ref{tab:ablation}, removing the strategy reward leads to a substantial performance drop across all benchmarks, with the most significant degradation observed on MVToolBench (from 60.1 to 50.7). This confirms that a simple outcome-based reward is insufficient for learning complex tool-use strategies. The dense process signals from the strategy reward are critical for guiding the model to discover and reinforce effective reasoning paths.

Similarly, removing the constraint penalties also degrades performance, particularly on V* and MVToolBench. This highlights the importance of penalizing inefficient or illogical actions. Without these penalties, the model is susceptible to "reward hacking," where it learns to maximize rewards through superfluous actions (\textit{e.g.,} rotating an already-correct image) that do not contribute to solving the task. The penalties serve as essential guardrails that promote more deliberate and efficient problem-solving. Figure~\ref{fig:ablation_reward_curves} visualizes this by showing that our full reward function leads to a more stable and higher final success rate during RL training compared to the ablated versions.

\begin{table}
    \centering
    \caption{Ablation study of our key reward components on the CodeVision-7B model. Removing either the strategy reward or the constraint penalties leads to a notable drop in performance.}
    \label{tab:ablation}
    \resizebox{0.6\textwidth}{!}{%
    \begin{tabular}{@{}l|cc|cc|c|c@{}}
    \toprule
    \multirow{2}{*}{Model} & \multicolumn{2}{c}{OCRBench} & \multicolumn{2}{c}{ChartQAPro} & \multirow{2}{*}{V*} & \multirow{2}{*}{MVToolBench} \\ \cmidrule(lr){2-5}
     & Rot180 & Verti & Rot180 & Hori &  &  \\ \midrule
    Qwen2.5-VL-7B & 70.2 & 17.0 & 23.4 & 19.5 & 74.6 & 18.1 \\
    Qwen2.5-VL-7B-SFT & 57.0 & 35.8 & 23.2 & 20.9 & 71.7 & 26.6 \\
    CodeVision-7B & 72.3 & 67.4 & 30.8 & 30.1 & 83.7 & 60.1 \\\midrule
    w/o Strategy Reward & 60.9 & 61.5 & 24.6 & 28.9 & 78.5 & 50.7 \\
    w/o Penalty & 68.3 & 66.3 & 24.0 & 24.3 & 71.2 & 55.9 \\ \bottomrule
    \end{tabular}%
    }
\end{table}

\begin{figure*}[h!]
    \centering
    \begin{subfigure}{0.245\textwidth}
        \centering
        \includegraphics[width=\textwidth]{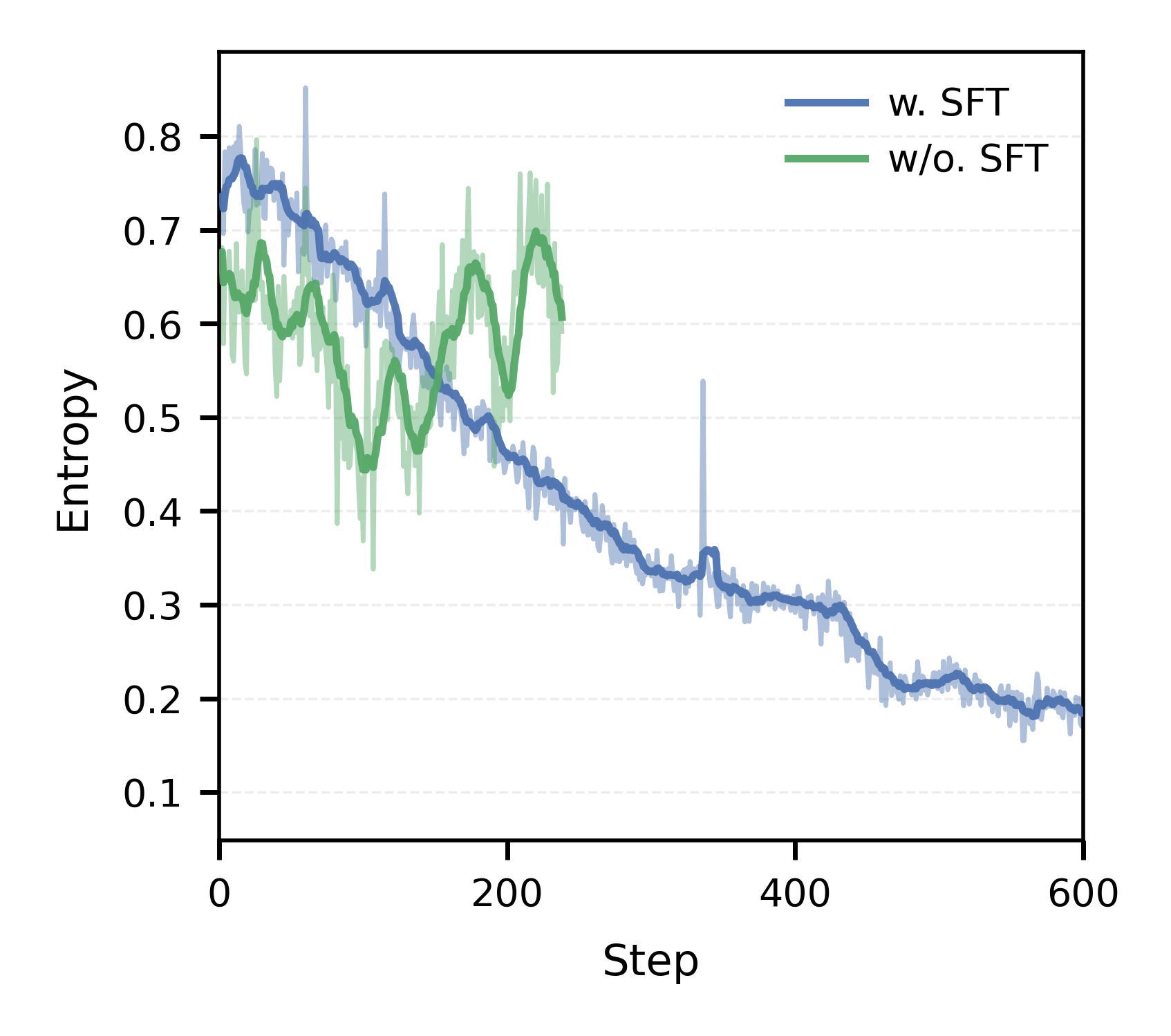}
        \caption{Entropy}
        \label{fig:sft_ablation:a}
    \end{subfigure}
    \hfill 
    \begin{subfigure}{0.245\textwidth}
        \centering
        \includegraphics[width=\textwidth]{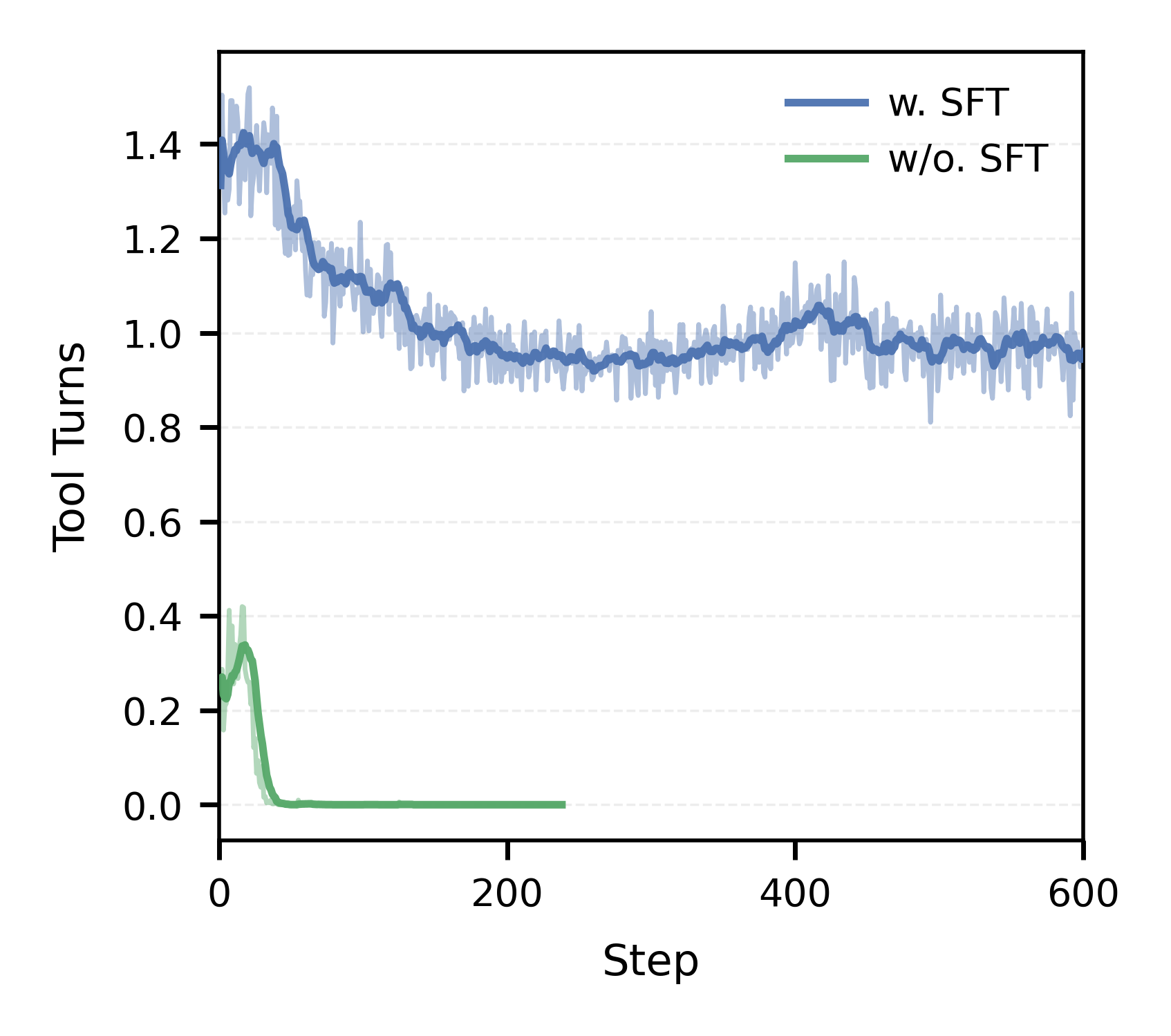}
        \caption{Tool Turns}
        \label{fig:sft_ablation:b}
    \end{subfigure}
    \hfill
    \begin{subfigure}{0.245\textwidth}
        \centering
        \includegraphics[width=\textwidth]{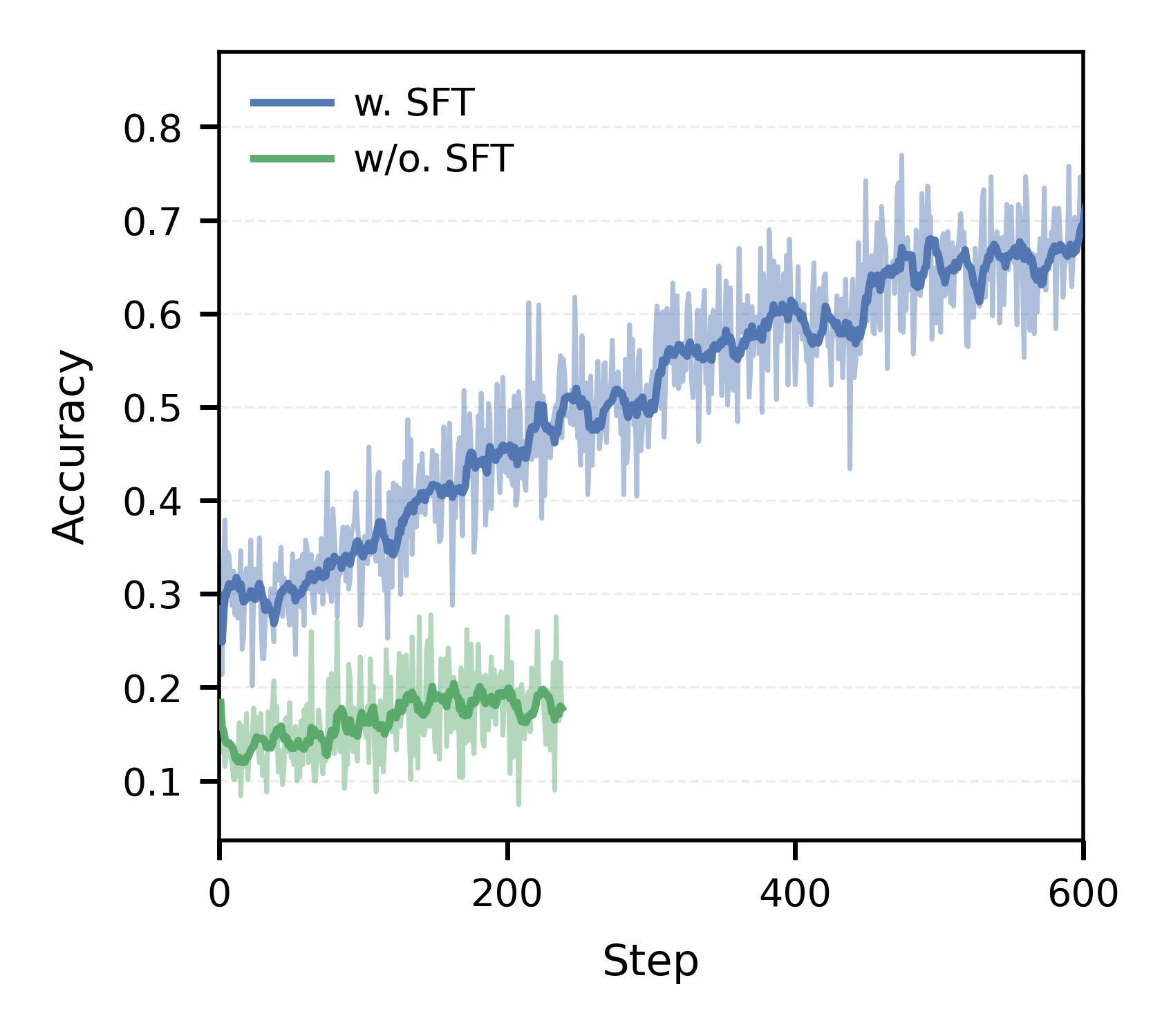}
        \caption{Accuracy Reward}
        \label{fig:sft_ablation:c}
    \end{subfigure}
    \hfill
    \begin{subfigure}{0.245\textwidth}
        \centering
        \includegraphics[width=\textwidth]{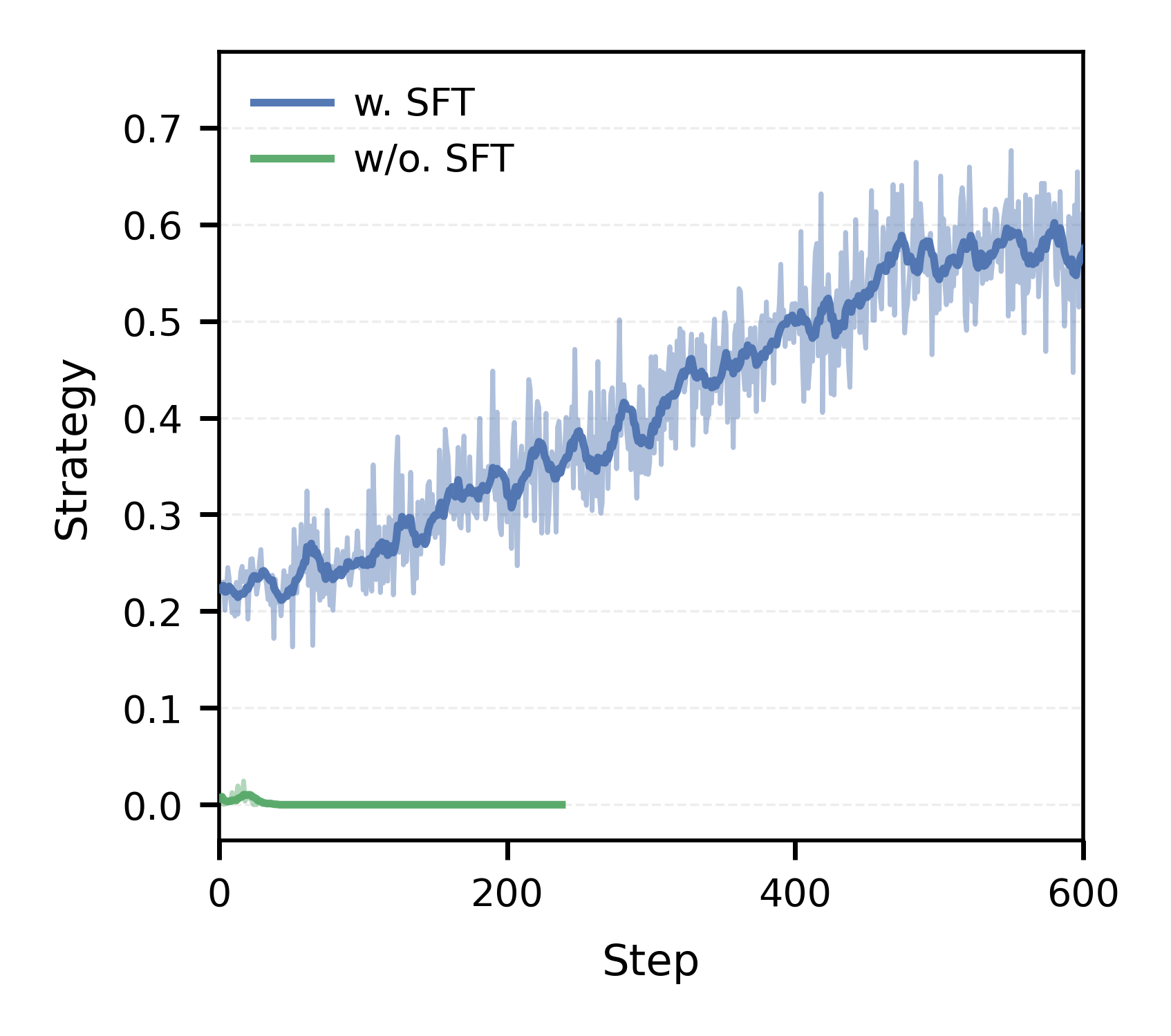}
        \caption{Strategy Reward}
        \label{fig:sft_ablation:d}
    \end{subfigure}
    \caption{Comparison of training with and without the SFT cold-start stage. Starting RL from a SFT-tuned checkpoint is critical for effective learning, while training from scratch fails to converge.}
    \label{fig:sft_ablation}
\end{figure*}

\textbf{Necessity of the cold start.}
We also study the importance of our two-stage training process. We compare our full model, which is first fine-tuned with SFT and then trained with RL, against a baseline trained with RL directly on the base model without the SFT cold start. As illustrated in Figure~\ref{fig:sft_ablation}, the model trained without SFT fails to achieve meaningful improvement. This is because the vast and unstructured action space of code generation makes it extremely difficult for the agent to discover useful tool-use policies through pure RL exploration. The SFT stage is crucial for bootstrapping the model, teaching it the fundamental syntax of tool calls and providing a strong initial policy. This warm start makes the subsequent RL phase significantly more stable and effective, allowing the model to refine its strategic reasoning on a solid foundation.

\section{Conclusion}
In this paper, we identify a critical vulnerability in state-of-the-art MLLMs: their pronounced brittleness to natural corruptions such as orientation changes. We introduce CodeVision, the ``code-as-tool'' framework, a flexible paradigm that empowers models to generate code for a virtually unlimited range of visual operations, moving beyond fixed toolsets. We train our models using a two-stage process combining SFT and RL with a dense, process-oriented reward function to foster strategic tool use. Experiments on our newly created benchmarks demonstrate that this approach remedies the identified weakness and significantly enhances model reasoning. Our framework fosters emergent behaviors like efficient tool-chaining and error recovery, paving the way for more capable MLLMs that treat visual interaction as a programming task.

\begin{figure*}
    \centering
    \begin{subfigure}{0.322\textwidth}
        \centering
        \includegraphics[width=\textwidth]{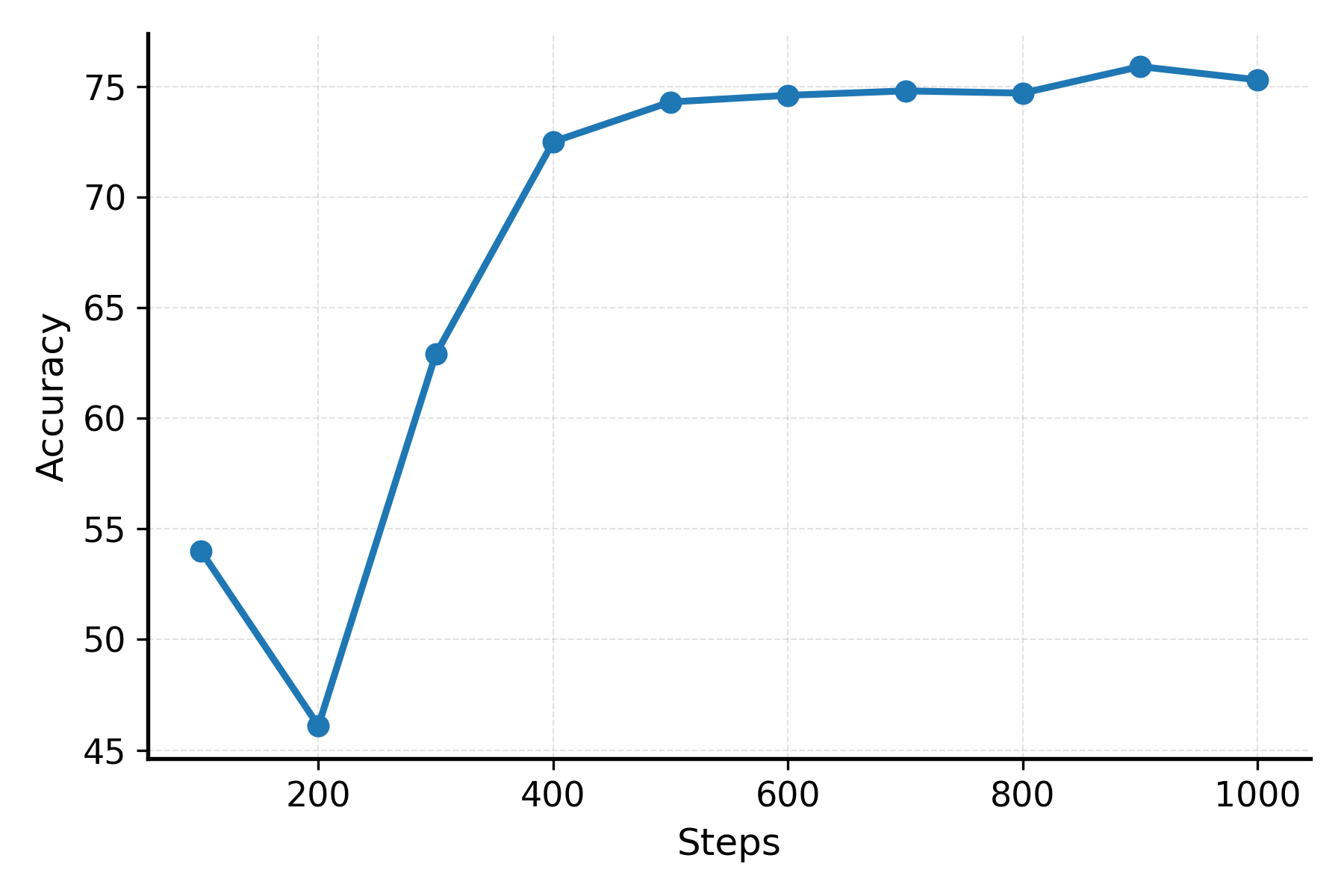}
        \caption{OCRBench-Rot90}
        \label{fig:metric_curve:a}
    \end{subfigure}
    \hfill 
    \begin{subfigure}{0.322\textwidth}
        \centering
        \includegraphics[width=\textwidth]{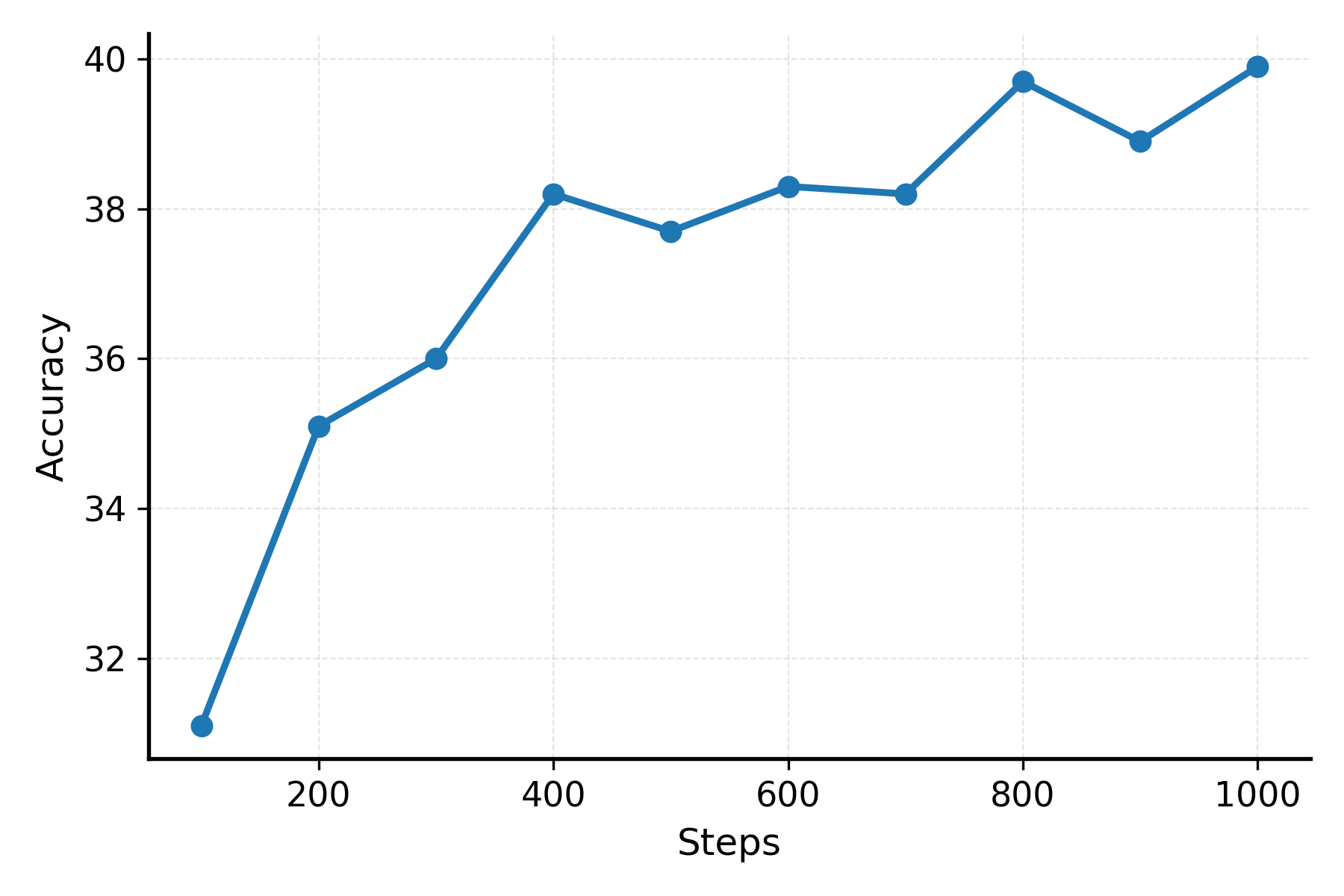}
        \caption{ChartQAPro-Hori}
        \label{fig:metric_curve:b}
    \end{subfigure}
    \hfill
    \begin{subfigure}{0.322\textwidth}
        \centering
        \includegraphics[width=\textwidth]{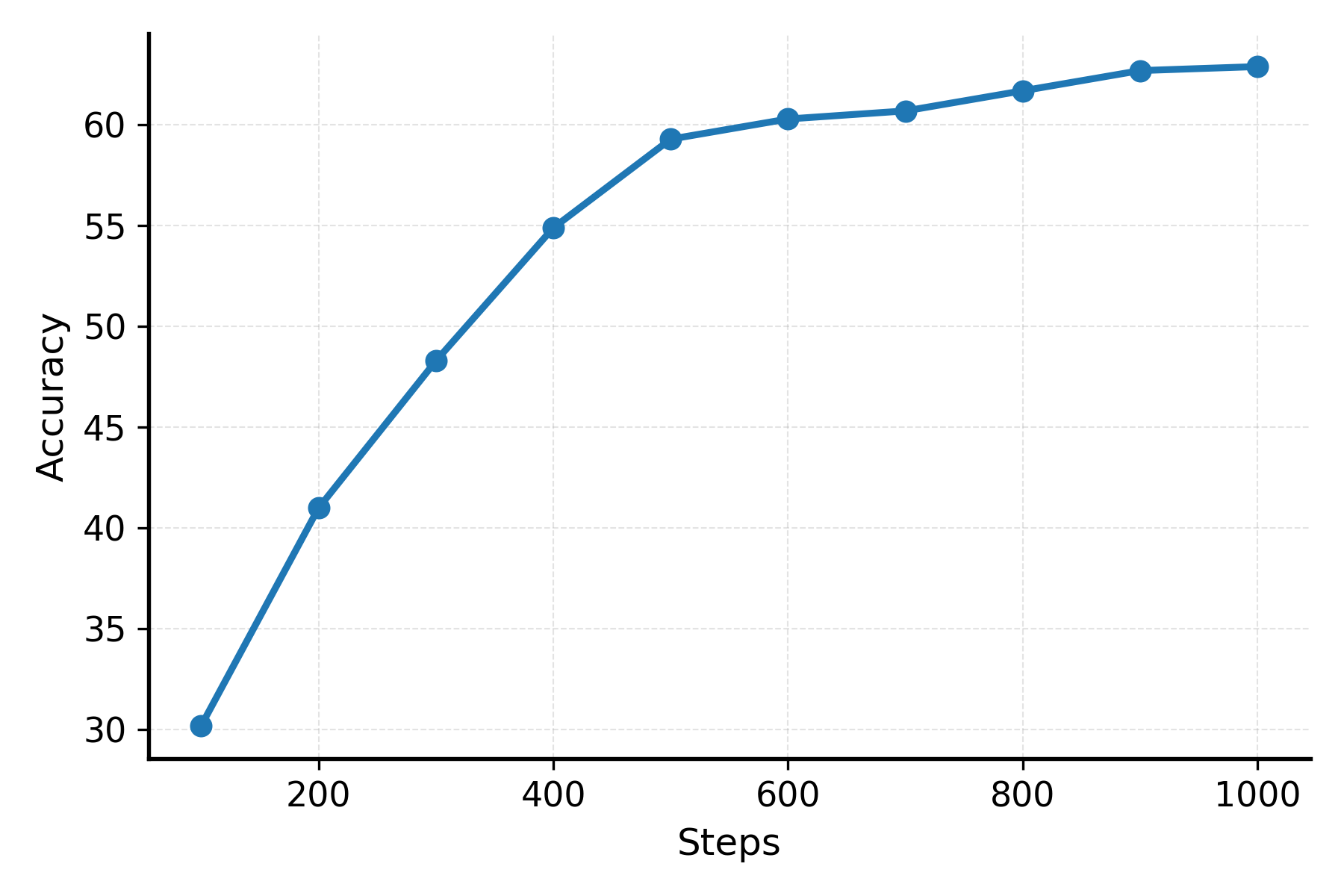}
        \caption{MVToolBench}
        \label{fig:metric_curve:c}
    \end{subfigure}
    \caption{Accuracy on OCRBench, ChartQAPro, and MVToolBench during RL training. The steady improvement across all benchmarks indicates that the model's learning has not yet saturated and could benefit from further scaling of data and tasks.}
    \label{fig:metric_curve}
\end{figure*}

\section{Limitations and Future Work}

In this work, we have focused on a core set of tools—primarily orientation correction and cropping—to establish the CodeVision framework and demonstrate its effectiveness in addressing model brittleness. While this targeted approach has proven successful, creating a truly general-purpose visual agent requires scaling and expansion in several key dimensions. Our work reveals several limitations that open up exciting avenues for future research:

\textbf{Expanding Tool Diversity and Compositionality.}
Our current research deliberately concentrates on a limited toolset. To enhance the model's generalization capabilities, future work should incorporate a much wider variety of tool types and data. This includes training on tasks that require more complex tool compositions and even multi-image tool use (\textit{e.g.,} comparing, merging, or analyzing multiple images simultaneously). Furthermore, the ``code-as-tool'' paradigm can be extended beyond standard Python libraries to include custom tools, such as proprietary search engines or generative models. By exposing only a simple API endpoint to the model, we can empower it to leverage powerful, black-box functionalities, further broadening its problem-solving horizons.

\textbf{Refining Process Supervision with Beneficial Tools.}
Our current training methodology relies on a ``must-use'' tool list to provide strong process supervision. While effective, this can be expanded to a more flexible framework. Future iterations could incorporate a broader set of ``beneficial'' tools. These are tools that are not strictly necessary to solve a task but are highly likely to improve performance, such as contrast enhancement for low-light images or a preliminary crop for focusing attention. By rewarding the use of such advantageous tools, we can encourage the model to learn more nuanced strategies and generalize better to tasks where the optimal toolset is not rigidly defined.

\textbf{Scaling Data, Tasks, and Model Exploration.}
Our experiments indicate that the model's learning has not yet reached saturation. The entropy of its policy suggests that there is still significant room for exploration and improvement. Moreover, as shown in Figure~\ref{fig:metric_curve}, we observe that performance on our benchmarks continues to increase steadily throughout training with no signs of plateauing. This strongly suggests that our approach would benefit from scaling up. By sourcing more diverse data, incorporating a wider array of tool types, and designing more varied tasks, we believe there is a clear path toward developing even more capable and robust visual agents.

\bibliographystyle{apalike}
\bibliography{ref}

\end{document}